\pdfoutput=1

\documentclass[11pt]{article}

\usepackage{acl}

\usepackage{times}
\usepackage{latexsym}

\usepackage[T1]{fontenc}

\usepackage[utf8]{inputenc}

\usepackage{microtype}

\usepackage[T1]{fontenc}
\usepackage[utf8]{inputenc}
\usepackage{microtype}
\usepackage{inconsolata}
\usepackage{multirow}
\usepackage{tabularx,ragged2e}

\usepackage{array,colortbl,xcolor}

\usepackage{graphicx} 

\usepackage{listings}

\usepackage{titlesec}

\newcommand{\citeg}[1]{\citep[e.g.,][]{#1}}

\title{Simple Linguistic Inferences of Large Language Models (LLMs):\\Blind Spots and Blinds}

\author{Victoria Basmov\textsuperscript{\normalfont1,2} \, Yoav Goldberg\textsuperscript{\normalfont1,2} \,Reut Tsarfaty\textsuperscript{\normalfont 1}\\
\textsuperscript{1}Bar-Ilan University \, \textsuperscript{2}Allen Institute for Artificial Intelligence \\ 
{\tt\{\href{mailto:vikasaeta@gmail.com}{vikasaeta}, \href{mailto:yoav.goldberg@gmail.com}{yoav.goldberg},
\href{mailto:reut.tsarfaty@gmail.com}{reut.tsarfaty}\}
@gmail.com}}

\begin{document}
\maketitle
\begin{abstract}
We evaluate LLMs' language understanding capacities on simple inference tasks that most humans find trivial. Specifically, we target (i) grammatically-specified entailments, (ii) premises with evidential adverbs of uncertainty, and (iii) monotonicity entailments. We design evaluation sets for these tasks and conduct experiments in both zero-shot and chain-of-thought setups, and with multiple prompts and LLMs. The models exhibit moderate to low performance on these evaluation sets.
Subsequent experiments show that embedding the premise in syntactic constructions that should preserve the entailment relations (presupposition triggers) or change them (non-factives), further confuses the models, causing them to either under-predict or over-predict certain entailment labels regardless of the true relation, and often disregarding the nature of the embedding context.
 Overall these results suggest that, despite LLMs' 
celebrated language understanding capacity, even the strongest models have blindspots with respect to certain types of entailments, and certain information-packaging structures act as ``blinds'' overshadowing the semantics of the embedded premise.

\end{abstract}

\section{Introduction}

LLMs have gained immense popularity thanks to their unprecedented ability to understand user queries and generate fluent seemingly-human responses. At the same time, people constantly report LLMs' failures, anecdotal
\citep{borji2023categorical} and systematic, e.g, the lack of reliability and consistency \citep{shen2023chatgpt, jang2023consistency,plevris2023chatbots},
contradictory or unreasonable answers \citep{zhong2023chatgpt}, inability to detect false assumptions \citep{shen2023chatgpt}, wrong information in prompts
\citep {zuccon2023dr}, contradictory responses to identical queries \citep{jang2023consistency, plevris2023chatbots}.

\begin{figure*}[t!]
\centering
\includegraphics[width=0.70\textwidth]{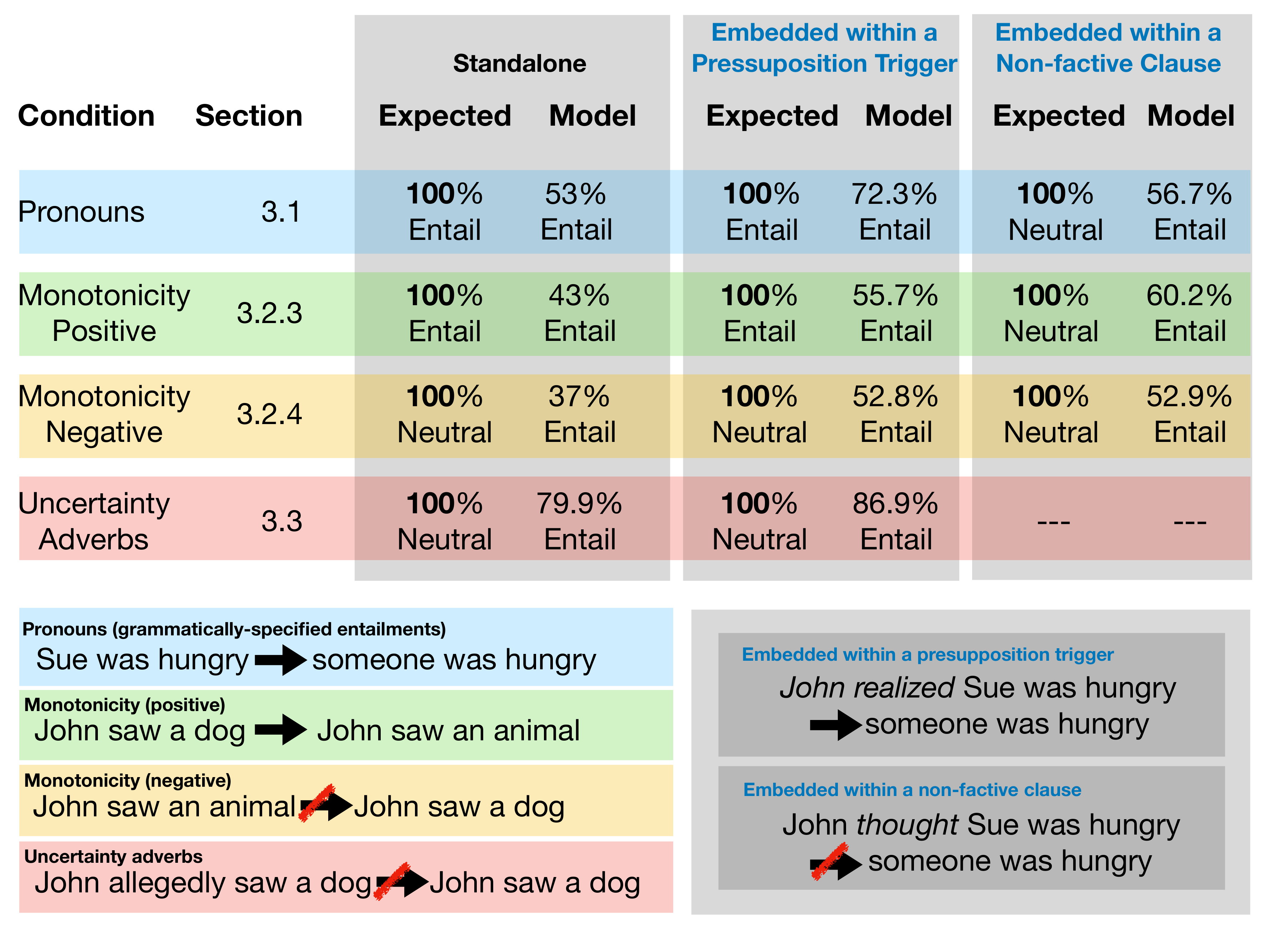}
\caption{High-level summary of the experiments and results (reported numbers are for gpt-3.5-turbo-0301).}
\label{fig:main}
\end{figure*}

However, humans are prone to some failures as well, e.g., overlooking false assumptions in questions beyond their area of expertise, or failing to find the correct solution to a math problem.

In this work we focus on tasks that are trivial for humans, and do not require any specialized expertise beyond proficiency in English.
For example, it is obvious to a human that \textit{Her brother was singing} entails \textit{Someone was singing}, and \textit {Fred’s tie is very long} implies \textit {Fred’s tie is long}, but not vice versa. However, as shall be seen shortly, 
LLMs fail to establish such systematic relations correctly. LLMs' errors on such simple tasks are much more indicative of absence of \emph{human-like} text understanding. 

We experiment with several types of natural language inferences (NLI), (a.k.a. recognizing textual entailment \cite{dagan2005pascal,bowman2015snli}), 
that are easy for humans, and show that they pose a challenge to LLMs. These NLI tests reveal some of the models' blind spots, and indicate that they are far from a genuine \emph{human-level} understanding. Moreover,
some information-packaging structures, where a statement is embedded within a surrounding linguistic context, may act as ``blinds'' that hinder the semantics of embedded premises, again in contrast to human-like behavior.

A summary of the conditions we test, as well as the trends observed for one model (gpt-3.5-turbo-0301) is available in Figure \ref{fig:main}.\footnote{Results for other models are presented in Table \ref{tab:all1} and discussed in Section \ref{experiments}. While directions and magnitudes differ, all models are far from human level, and do not respond correctly to embedding environments.}

We focus on inference types that are solely based on common linguistic phenomena and ``trival'' world-knowledge such as class membership (``a dog is an animal'', ``navy blue is a shade of blue''). Specifically, we test LLMs' ability to make three inference types:
(i) {\em Grammatically-specified entailments}, i.e. replacing a constituent of the premise with an indefinite pronoun as \emph{somebody} or \emph{something}.
(ii) Premises with {\em evidential adverbs of uncertainty} (\textit{supposedly}, \textit{allegedly} etc.), that
block the entailment of the rest of the clause, and
(iii) {\em Monotonicity entailment} (see \citet{maccartney-manning-2008-modeling}) of two kinds: upward , i.e. from subsets to supersets (``Jack is a dog'' entails ``Jack is an animal''), and downward, i.e. from supersets to subsets (``Jack isn’t an animal'' entails ``Jack isn’t a dog'').
We manually curate test sets for these inference types and experiment with them in a zero-shot setup, observing that LLMs struggle with these phenomena, leading to low accuracy.

We next check how embedding of the premise in a larger grammatical context affects the prediction. Such embedding can take several forms. Contexts consisting of presupposition triggers (e.g. \textit{He realized that [...], They were glad that [...], Something happened before [...]}) serve to \emph{strengthen} the embedded premise, while similarly structured non-factives (e.g. \textit{I feel that [...],
He imagined that [...]}) may \emph{cancel} it. We experiment with both context types and show that in most cases they affect the LLMs' predictions incorrectly. E.g., ChatGPT\footnote{\url{https://openai.com/chatgpt}} has a hard time discerning the two cases,
incorrectly treating both as hints towards entailment (for regular prompting) or against it (for chain-of-thought prompting).
These or similar trends are observed across different prompts and models (GPT-3.5, GPT-4, LLaMA 2). While GPT-4 clearly improves over other LLMs in terms of accuracy and shows some more sound trends, it still leaves much room for improvement compared to the human baseline.

These results demonstrate that state-of-the-art LLMs were unable to learn simple linguistic inferences that humans find trivial: they did not acquire them automatically in pre-training, and also in the process of instruct-tuning or human-feedback tuning.
Persistence of the problem across prompts and LLMs implies that this is a systematic issue.

While current LLMs struggle with these trivial phenomena, we hope future models will show improvements. Our methodology and the dataset we release\footnote{URL will be provided in camera ready} can help track this progress.

\section{Linguistic phenomena considered}
\label{ling_back}
We focus on the following linguistic phenomena.
\paragraph{Gramatically-specified entailments}
The set of the entailments of any sentence includes so-called \textit{grammatically-specified entailments} \citep{Wilson79}, i.e., 
entailments where a constituent of the premise is substituted with a variable (such as an indefinite pronoun like \textit{somebody}, \textit{something} etc.).
For instance, the entailments of \textit{“You’ve eaten all my apples”} include, among others:
\begin{quote} 
\textit{You've eaten all someone's apples.}

\textit{You've eaten all of something.}

\textit{You've eaten something.}

\textit{You've done something.}

\textit{Someone's eaten all my apples.}
\end{quote}
\paragraph{Monotonicity entailments}
hold when less specific predicates are substituted with more specific ones, or vice versa. 
They can be of two types:
\begin{itemize}
 
 \item Upward: more specific predicates can be substituted with less specific ones.
 \begin{quote}
 \textit{Jack is a dog. $\models$ Jack is an animal.} 
 \end{quote}

 \item Downward: less specific predicates can be substituted with more specific ones.
 \begin{quote}
 \textit{All animals need water. $\models$ All dogs need water.}
 \end{quote}

\end{itemize}

\paragraph{Evidential Adverbs}
``express degrees of certitude with respect to the speaker’s
subjective perception of the truth of a proposition'' \citep{advs_book}.
We
test LLMs' ability to understand evidential adverbs expressing \emph{uncertainty} (\textit{allegedly}, \textit{purportedly}, \textit{supposedly} etc.).
Introducing such adverbs into a clause cancels the entailment of the rest of the clause. E.g., \textit{Mike allegedly worked all night} does not entail
\textit{Mike indeed worked all night}. 
The relation between the two statements is neutral.

\paragraph{Presuppositions and Presupposition Triggers}
\textit{Presupposition} \citep{sep-presupposition,jeretic-etal-2020-natural,parrish-etal-2021-nope} is a type of inference “whose truth is taken for granted in the utterance of a sentence” \citep{Huang2011}.
Below, \textbf{a} presupposes \textbf{b} (i.e. if \textbf{b} is false, \textbf{a} cannot be felicitously
uttered):
\begin{quote}
\textbf{a}. \textit{Jane returned to New York.}

$\models$ \textbf{b}. \textit{Jane has been to New York before.} 
\end{quote}

Presuppositions are not presented as at-issue content of the utterance, but rather as part of the background,
mutually known or assumed by the speaker and the hearer (even if in reality it is not the case). The speaker of \textbf{a} does not \textit{inform} the hearer that Jane has been to New York before:
she \textit{assumes} it; and if the hearer does not know it, she \textit{accommodates} it upon hearing the utterance \citep{presup_accommodation}.

Presuppositions are normally evoked by constructions or lexical items, called \textit{presupposition triggers} \citep{Karttunen2016PresuppositionWW}. In sentence \textbf{a} above, the presupposition is triggered by the verb \textit{returned}, from the class of \textit{iterative verbs} which presuppose that the action has happened before. Other iterative verbs are \textit{relearn, reread, reapply} etc.
Presupposition triggers used in this work are factives, temporal and other adverbial clauses and embedded \textit{wh}-questions. 

\paragraph{Non-factive Verbs and Expressions}
\citep{KiparskyKiparsky+1970+143+173}, such as \textit{believe, claim, feel,
hope,
suspect, think}, \textbf{do not entail either truth or falsity of their complements}. For example, given:
\begin{quote}
\textbf{a}. \textit{Jane thinks that Bill bought bread.}

\textbf{b}. \textit{Bill bought bread.}

\textbf{c}. \textit{Bill didn't buy bread.} 
\end{quote}
Sentence \textbf{a} does not entail either \textbf{b} or \textbf{c}. The relation between \textbf{a} and \textbf{b} is neutral, and so is the relation between \textbf{a} and \textbf{c}. 

\paragraph{Presupposition Triggers, Non-Factives and NLI}
It is important to note that embedding a premise under a presupposition trigger does not affect the relations between the premise and hypothesis. By contrast, if we embed the premise under a non-factive, the relation becomes neutral.
For example:
\begin{quote}
\textbf{a}. \textit{A balloon hit a light post.}

 $\models$ \textbf{b}. \textit{Something hit a light post} 

\end{quote}
Premise \textbf{a} above entails hypothesis \textbf{b}.
If we embed premise \textbf{b} under a presupposition trigger as in {\textbf{a'}: 
\begin{quote}
\textit{\textbf{a'}. \underline{She realized that} a balloon hit a light post}

$\models$ \textbf{b}. \textit{Something hit a light post} 
\end{quote}
the relation does not change: the new premise \textbf{a'} still entails \textbf{b}.
However, when embedding premise \textbf{a} under a non-factive verb:
\begin{quote}
\textit{\textbf{a''}. \underline{I suspect} a balloon hit a light post}

$\not\models$ \textbf{b}. \textit{Something hit a light post} 
\end{quote}
the relation becomes neutral: without additional context the new premise \textbf{a''} does not entail \textbf{b}.
\section{Main Experiments}\label{experiments}

\paragraph{Benchmark Design.} We construct our benchmark in the form of several groups of NLI pairs, where each group focuses on a single linguistic phenomenon, and all the items within a group share the same label. This allows us to isolate the models' performance on the different conditions, rather than just providing aggregate accuracy numbers, which hide the underlying linguistic trends. 

\paragraph{Zero-shot Setup.}
We deliberately choose a zero-shot setup rather than in-context learning, for the following reasons. 1) All tested patterns are easily learnable from examples due to their consistent structure within each group, so predictions based on such learning would be a form of “parroting” rather than “understanding”. 2) It is zero-shot capabilities that indicate if an LLM is truly a general-purpose system \citep{qin2023chatgpt}. In real-world scenarios, the goal is not necessarily to handle these inferences directly, but rather to use them as a means to an end. Hence, zero-shot capability becomes crucial. 3) Finally and most importantly, our goal was not to answer “can we make an LLM perform well on these simple cases through prompting” but rather to answer “is this semantic picked up by an LLM automatically as part of its training, from text alone'' as some works seemingly suggest \citeg{merrill-etal-2022-entailment}. 

\paragraph{Models.} We test a range of OpenAI models: GPT-3.5 (text-davinci-003), ChatGPT (gpt-3.5-turbo-0301 and gpt-3.5-turbo-0613), and GPT4 (gpt-4-0314 and gpt-4-0613). 
We access these LLMs through OpenAI's API\footnote{\url{https://openai.com/product}} with the default settings\footnote{For text-davinci-003 we set the temperature at 0.},
using a single prompt which receives two texts and asks if, given text 1, text 2 is true, false or neutral (see Appendix \ref{sec:orig_prompt} for details).\footnote{This prompt yields a 71\% accuracy on 300-instance sample from the SNLI dataset, consistent with
previous SNLI prompting literature \citep{qin2023chatgpt,wang2023robustness,jang2023consistency}.}

We also evaluate the 70B LLaMA-2 Chat model \citep{touvron2023LLaMA}.\footnote{We access LLaMA 2 through the Replicate API: \url{https://replicate.com}. Since the prompt used for the OpenAI models yields very poor accuracy (39\%) on SNLI for LLaMA 2, we used a different prompt template, with SNLI accuracy of 61\%. We use temperature of 0.01 and top-k=1.}

\paragraph{Data Collection Methodology.}
For monotonicity entailment, we used a sample from the MED dataset \citep{yanaka-etal-2019-neural} (see \S\ref{monotonicity_p}, \S\ref{monotonicity_n}).
For grammatically-specified entailments and uncertainty adverbs, the data was created by the first author by manually collecting naturally-occurring sentences from web-pages retrieved by random keywords and selecting a subset of sentences in the page that fit our desired criteria (e.g., for grammatically-specified entailment - sentences containing noun phrases in different positions that could be substituted by indefinite pronouns), and potentially re-writing them (for example, removing embedding contexts) to obtain simple sentences. We did not consult an LLM during this process. We then derived premise-hypotheses pairs from these sentences as described in \S\ref{gse} and \S\ref{advs}. For the embedding contexts, we manually curated a set of candidate environments by collecting naturally occurring presupposition triggers and non-factives from textual material on the same web-pages.

The labels were assigned based on the linguistic properties of the phenomena (e.g. entailment-cancelling property of uncertainty adverbs).
The resulting labels for each data group are confirmed by the majority vote in our human annotation experiment (see Section \ref{human}) in 99.8\% of the cases and by all three annotators unanimously in 95.8\% of the cases\footnote{These numbers differ from the ones reported in \S\ref{human}, because here we do not consider the monotonicity data, which was taken from an existing dataset.} with a fleiss kappa of 0.94, which corresponds to almost perfect agreement, attesting to the high quality of the data.

\begin{table*}[t!]

\resizebox{\textwidth}{!}{%
\begin{tabular}{|
>{\columncolor[HTML]{D9D9D9}}l |l|l|l|}
\hline
\textbf{Inference type} &
  \cellcolor[HTML]{D9D9D9}\textbf{Standalone} &
  \cellcolor[HTML]{D9D9D9}\textbf{\begin{tabular}[c]{@{}l@{}}Under presupposition triggers\\(premise modified;\\label unmodified)\end{tabular}} &
  \cellcolor[HTML]{D9D9D9}\textbf{\begin{tabular}[c]{@{}l@{}}Under non-factives\\(premise modified;\\label - always neutral)\end{tabular}}\\\hline
\textbf{\begin{tabular}[c]{@{}l@{}}Grammatically-\\specified\\entailments\end{tabular}}&
          \textit{\begin{tabular}[c]{@{}l@{}}\textup{(1) \textbf{P:}} Crown Princess Mary\\of Denmark has given birth\\to a healthy baby boy.\\\textbf{\textup{H:}} Someone has given\\birth to a healthy baby boy.\\\textup{\textbf{L:} entailment} \end{tabular}} &
  \textit{\begin{tabular}[c]{@{}l@{}}\textup{(5) \textbf{P:}} \textbf{We are happy that} Crown\\Princess Mary of Denmark has\\given birth to a healthy baby boy.\\\textbf{\textup{H:}} Someone has given\\birth to a healthy baby boy.\\\textup{\textbf{L:} entailment}\end{tabular}} &
  \textit{\begin{tabular}[c]{@{}l@{}}\textup{(9) \textbf{P:}} \textbf{I hope} Crown\\Princess Mary of Denmark has\\given birth to a healthy baby boy.\\\textbf{\textup{H:}} Someone has given\\birth to a healthy baby boy.\\\textup{\textbf{L:} neutral}\end{tabular}}\\\hline
\textbf{\begin{tabular}[c]{@{}l@{}}Monotonicity\\positive\end{tabular}} &
  \textit{\begin{tabular}[c]{@{}l@{}}\textup{(2)\textbf{ P:}} She planted  blue\\and purple pansies in\\the flower bed.\\\textbf{\textup{H:}} She planted pansies\\in the flower bed.\\\textup{\textbf{L:} entailment}\end{tabular}} &
  \textit{\begin{tabular}[c]{@{}l@{}}\textup{(6) \textbf{P:}} \textbf{After} she planted  blue and\\purple pansies in the flower bed,\\she started planting other flowers.\\\textbf{\textup{H:}} She planted pansies\\in the flower bed.\\\textup{\textbf{L:} entailment}\end{tabular}} &
  \textit{\begin{tabular}[c]{@{}l@{}}\textup{(10) \textbf{P:}} \textbf{I think} she planted  blue\\and purple pansies in\\the flower bed.\\\textbf{\textup{H:}} She planted pansies\\in the flower bed.\\\textup{\textbf{L:} neutral}\end{tabular}}\\\hline
\textbf{\begin{tabular}[c]{@{}l@{}}Monotonicity\\negative\end{tabular}} &
  \textit{\begin{tabular}[c]{@{}l@{}}\textup{(3) \textbf{P:}} Susan made\\a dress for Jill.\\\textbf{\textup{H:}} Susan made\\a long dress for Jill.\\\textup{\textbf{L:} neutral}\end{tabular}} &
  \textit{\begin{tabular}[c]{@{}l@{}}\textup{(7) \textbf{P:}} \textbf{They are aware that}\\Susan made a dress for Jill.\\\textbf{\textup{H:}} Susan made\\a long dress for Jill.\\\textup{\textbf{L:} neutral}\end{tabular}} &
  \textit{\begin{tabular}[c]{@{}l@{}}\textup{(11)\textbf{ P:}} \textbf{They believe}\\Susan made a dress for Jill.\\\textbf{\textup{H:}} Susan made\\a long dress for Jill.\\\textup{\textbf{L:} neutral}\end{tabular}}\\\hline
\textbf{Adverbs} &
  \textit{\begin{tabular}[c]{@{}l@{}}\textup{(4) \textbf{P:}} These persons\\were allegedly inhabiting\\the home.\\\textbf{\textup{H:}} These persons were\\ inhabiting the home.\\\textup{\textbf{L:} neutral}\end{tabular}} &
   \textit{\begin{tabular}[c]{@{}l@{}}\textup{(8) \textbf{P:}} \textbf{The owner was aware}\\\textbf{that} these persons were\\ allegedly inhabiting the home.\\\textbf{\textup{H:}} These persons were\\inhabiting the home.\\\textup{\textbf{L:} neutral}\end{tabular}} &
  \textit{N/A}\\\hline
\end{tabular}
}

\caption{Inference types (P - premise, H - hypothesis, L - label) }
\label{tab:inf_types1}
\end{table*}

\subsection{Human Performance} \label{human} To verify that the inference types we consider are indeed trivial for humans, we sample 100 examples from each of the 11 data types detailed in Figure \ref{fig:main} 
and have each example labeled by three human annotators from Amazon Mechanical Turk,
using majority vote to obtain the final label.
We obtain an overall accuracy of 97.55\%, and a full agreement among the annotators for 91\% of the examples. These very high numbers confirm that these inference types pose no challenge for humans. For results grouped by inference type see Table \ref{tab:all1}. More details are available in Appendix \ref{sec:h_baseline}.

\subsection {Testing LLMs in an NLI setting} \label{standalone}

Table \ref{tab:all1} (top) summarizes the results.\footnote{All the results reported in this paper are from a single run.} For brevity, we include only one ChatGPT and one GPT-4 version in the table. Results for the other versions can be found in Appendix \ref{sec:acc}.

\subsubsection{Grammatically-Specified Entailments}\label{gse}

\paragraph{Data:} 
We manually curated a dataset of 100 pairs with grammatically specified entailments 
replacing an NP in each premise with an indefinite pronoun in the hypothesis (see Table \ref{tab:inf_types1}, example (1)).
This is a seemingly very easy dataset, trivial for any human, where all gold labels are ``ENTAILMENT''.

\paragraph{Results:} All LLMs except GPT4 perform poorly on these items (maximum 53\% accuracy). GPT-4 versions handle them much better, with 84\% and 100\% (human-level) accuracy respectively.
\subsubsection{Positive Monotonicity Entailment} \label{monotonicity_p}
\paragraph{Data:} We sample 100 positive examples (the gold label is “ENTAILMENT”) from the Monotonicity Entailment Dataset (MED) \citep{yanaka-etal-2019-neural} - see example (2) in Table \ref{tab:inf_types1}.

\noindent\textbf{Results:} Except for GPT-4, all models yield very low accuracy (far below 50). The highest result (by gpt-4-0613) of 70\% accuracy is still far below human performance. 

\subsubsection{Negative Monotonicity Entailment} \label{monotonicity_n}
\paragraph{Data:} We sample 100 negative examples (the gold label is “NEUTRAL”) from MED \citep{yanaka-etal-2019-neural} - see example (3) in Table \ref{tab:inf_types1}.

\noindent\textbf{Results:} 
Here, the \emph{earlier} GPT-4 version (gpt-4-0314) only slightly underperforms humans (88\%).
All other models (including gpt-4-0613) have much lower accuracy: 28\%- 56\%.\footnote{Similar results on MED are reported by \citet{liu2023evaluating}.}

\subsubsection{Adverbs of Uncertainty}\label{advs}
\label{advs_orig}
\paragraph{Data:} We manually create a dataset of 100 sentence pairs where
the premise contains an uncertainty adverb, while the hypothesis omits it (example (4) in Table \ref{tab:inf_types1}).
We apply 9 adverbs (\textit{allegedly, hopefully, possibly, presumably, probably, purportedly, reportedly, seemingly, supposedly}) to each of the 100 pairs, obtaining 900 examples, 100 per adverb. The gold label for all pairs is “NEUTRAL”.

\noindent\textbf{Results:}
Strongest result (70\% accuracy) is obtained by the earlier GPT-4 (0314), and is substantially below human performance. All other models, including the newer GPT-4, achieve very low accuracy on this data (4.67\% - 38\%). 

\noindent\textbf{Bottomline:} Except for GPT-4, all tested models perform very poorly on all the explored inference types. In many cases the LLMs (especially LLaMA 2) predict mostly \textit{the opposite of the expected label} (``ENTAILMENT'' where ``NEUTRAL'' is expected and vice versa). While the two GPT-4 versions improve significantly over other LLMs, neither handles all the inference types equally well: the later version (gpt-4-0613) is very good at handling the cases where ``ENTAILMENT'' is expected, but its predecessor (gpt-4-0314) outperforms it on ``NEUTRAL'' cases.
\subsection{Embedding the Premises under Presupposition Triggers} 
 Table \ref{tab:all1} (middle) summarizes the results (except for gpt-3.5-turbo-0613 and gpt-4-0314 covered in Appendix \ref{sec:acc}). 

\paragraph{Data:} Next we modify all the datasets described above by embedding the original premises under presupposition triggers (see examples (5)--(8) in Table \ref{tab:inf_types1}). We use 23 trigger types: 21 factives, such as \textit{know, realize, be glad}; embedded wh-questions, such as \textit{This explains \underline{why he came}}; and adverbial clauses, e.g. \textit{\underline{After he came}, I cooked dinner}. 

For uncertainty adverbs, we randomly sample 100 examples from our dataset of 900 sentence pairs, and apply each of the 23 triggers to each sampled pair, obtaining 2300 pairs. For other data types (grammatically specified entailments, positive and negative monotonicity entailments) we embed the 100 original premises under 23 triggers, which also results in 2300 pairs per data type. The hypotheses remain unchanged (see Table \ref{tab:inf_types1}).\footnote{In experiments with both embedding context types we use full data for text-davinci-003, gpt-3.5-turbo-0301 and LLaMA 2, and 100 examples per inference type for other models.}

Presupposition triggers do not change the relation between the premise and the hypothesis, so the gold label remains unchanged: ``ENTAILMENT'' for grammatically specified and positive monotonicity entailments; ``NEUTRAL'' for negative monotonicity entailments and uncertainty adverbs.

\paragraph{Results:}

The OpenAI models show the same pattern across all data types: under presupposition triggers they predict ``ENTAILMENT'' more \textit{regardless of the correct label}.
This improves accuracy where ``ENTAILMENT'' is expected (grammatically specified and positive monotonicity entailments), but reduces accuracy for ``NEUTRAL'' data (negative monotonicity, uncertainty adverbs).\footnote{The only exception is gpt-3.5-turbo-0613: under presupposition triggers it predicts less entailment (except for negative monotonicity), which in most cases hurts the accuracy (see Appendix \ref{sec:acc}).}

LLaMA 2 shows a ``healthier'' trend: under presupposition triggers it seems to capture all inference types better than in the standalone experiments, predicting more entailment for the ``positive'' datasets and more ``neutral'' relations for the ``neutral'' datasets. However, its accuracy remains consistently low (below 40\%) across all data types. Additionally, under presupposition triggers, it consistently predicts more ``contradiction'' labels for all inference types, which is always wrong.

\noindent\textbf{Bottomline:} Under presupposition triggers the OpenAI LLMs tend to predict more entailment irrespective of the correct label. LLaMA 2 improves in this setting, but shows a different wrong trend: predicting more ``contradiction'' labels in all cases.

\subsection{Embedding the Premises under Non-Factives} 
Table \ref{tab:all1} (bottom) summarizes the results (gpt-3.5-turbo-0613 and gpt-4-0314 are covered in Appendix \ref{sec:acc}). 
\paragraph{Data:} 
We modify the original datasets (see \S\ref{standalone}), embedding the premises under non-factives (see examples (9)--(11) in Table \ref{tab:inf_types1}). For grammatically-specified entailments and both types of monotonicity data, we embed each of the 100 original premises under 23 different non-factives, e.g. \textit{feel, hope, believe} etc., obtaining 2300 pairs for each data type. We omit this experiment for adverbs for semantic reasons: including both an uncertainty adverb and a non-factive into the premise (\textit{\underline{I guess} he \underline{allegedly} worked all night.}) results in double expression of uncertainty, creating a tautology.

The hypotheses remain unchanged (see Table \ref{tab:inf_types1}).

Since non-factives cancel entailment, the gold label in all the cases changes to ``NEUTRAL''.

\paragraph{Results:}
Surprisingly, GPT-3.5 and ChatGPT\footnote{The newer ChatGPT version, gpt-3.5-turbo-0601, shows the same trends as under presupposition triggers, predicting less entailment (except for negative monotonicity) (see Appendix \ref{sec:acc}). It shows that, just like its predecessors, this LLM is unable to distinguish the two types of embedding contexts. } again predict ``ENTAILMENT'' more often for all the data groups.\footnote{This inclination to overpredict entailment might imply the use of lexical overlap heuristics, given the lexical similarity between premises and hypotheses in our data. However, a closer examination reveals a contrary trend: standalone premises have greater lexical overlap with hypotheses than embedded ones, yet the models predict more entailment for embedded premises. Our evaluation on a subset of HANS \citep{mccoy-etal-2019-right} also suggest that lexical overlap does not significantly influence ChatGPT towards entailment.}
This hurts the accuracy for all data groups, since now the correct label is always “NEUTRAL”.
 Both GPT-4 versions show the \textit{correct} trend predicting \textit{less} ``ENTAILMENT'' and \textit{more} ``NEUTRAL'' labels under non-factives. Yet, the now-incorrect ``ENTAILMENT'' label remains very frequent reducing the accuracy. 
 
 LLaMA 2 again (as with presupposition triggers) seems more accurate compared to standalone experiments, but a closer look reveals it is likely ``right for the wrong reason'': the proportion of neutral predictions has changed very little compared to the standalone experiments (maximum by 7\%), but now the ``NEUTRAL'' label is always correct, which seemingly boosts the accuracy.

\subsubsection{Bottomline} Only the GPT-4 models capture the overall trend correctly reducing entailment predictions and increasing neutral predictions under non-factives. However, while, ideally, the entailment predictions should be now reduced to 0, they still remain quite frequent, especially in the supposedly more advanced gpt-4-0613, reducing the models' accuracy.

 Across all the experiments described above, GPT-4 models show the best results and the most human-like trends. They are also the only ones consistantly refraining from the always incorrect ``CONTRADICTION'' label. However, the earlier version (see Appendix \ref{sec:acc}) proves biased towards neutral predictions, while the later one constantly leans towards entailment. Under presupposition triggers they share their predecessors' erroneous trend: predicting more entailment regardless of the correct label. 
 In spite of the favourable dynamics both GPT-4 versions still perform far below the human level on most inference types we explored.
 
Notably, overall, the newer ChatGPT and GPT-4 versions prove less accurate than the respective earlier ones (see Appendix \ref{sec:acc}).


\begin{table*}[t!]
\resizebox{\textwidth}{!}{%

\begin{tabular}{|l|l|lll|lll|lll|lll|lll|}
\hline
 &
   &
  \multicolumn{3}{l|}{\textbf{human}} &
  \multicolumn{3}{l|}{\textbf{GPT-3.5}} &
  \multicolumn{3}{l|}{\textbf{ChatGPT}} &
  \multicolumn{3}{l|}{\textbf{GPT-4}} &
  \multicolumn{3}{l|}{\textbf{LLaMA 2 70b}} \\ \hline
 &
   &
  \multicolumn{1}{l|}{\textbf{e}} &
  \multicolumn{1}{l|}{\textbf{n}} &
  \textbf{c} &
  \multicolumn{1}{l|}{\textbf{e}} &
  \multicolumn{1}{l|}{\textbf{n}} &
  \textbf{c} &
  \multicolumn{1}{l|}{\textbf{e}} &
  \multicolumn{1}{l|}{\textbf{n}} &
  \textbf{c} &
  \multicolumn{1}{l|}{\textbf{e}} &
  \multicolumn{1}{l|}{\textbf{n}} &
  \textbf{c} &
  \multicolumn{1}{l|}{\textbf{e}} &
  \multicolumn{1}{l|}{\textbf{n}} &
  \textbf{c} \\ \hline
 &
  \cellcolor[HTML]{D9EAD3}\textbf{pronouns} &
  \multicolumn{1}{l|}{\cellcolor[HTML]{D9EAD3}\textbf{100.0}} &
  \multicolumn{1}{l|}{\cellcolor[HTML]{D9EAD3}0.0} &
  \cellcolor[HTML]{D9EAD3}0.0 &
  \multicolumn{1}{l|}{\cellcolor[HTML]{D9EAD3}\textbf{39.0}} &
  \multicolumn{1}{l|}{\cellcolor[HTML]{D9EAD3}56.0} &
  \cellcolor[HTML]{D9EAD3}5.0 &
  \multicolumn{1}{l|}{\cellcolor[HTML]{D9EAD3}\textbf{53.0}} &
  \multicolumn{1}{l|}{\cellcolor[HTML]{D9EAD3}34.0} &
  \cellcolor[HTML]{D9EAD3}13.0 &
  \multicolumn{1}{l|}{\cellcolor[HTML]{D9EAD3}{\underline{\textbf{100.0}}}} &
  \multicolumn{1}{l|}{\cellcolor[HTML]{D9EAD3}0.0} &
  \cellcolor[HTML]{D9EAD3}0.0 &
  \multicolumn{1}{l|}{\cellcolor[HTML]{D9EAD3}\textbf{31.0}} &
  \multicolumn{1}{l|}{\cellcolor[HTML]{D9EAD3}61.0} &
  \cellcolor[HTML]{D9EAD3}8.0 \\ \cline{2-17} 
 &
  \cellcolor[HTML]{D9EAD3}\textbf{monotonicity positives} &
  \multicolumn{1}{l|}{\cellcolor[HTML]{D9EAD3}\textbf{89.0}} &
  \multicolumn{1}{l|}{\cellcolor[HTML]{D9EAD3}11.0} &
  \cellcolor[HTML]{D9EAD3}0.0 &
  \multicolumn{1}{l|}{\cellcolor[HTML]{D9EAD3}\textbf{25.0}} &
  \multicolumn{1}{l|}{\cellcolor[HTML]{D9EAD3}46.0} &
  \cellcolor[HTML]{D9EAD3}29.0 &
  \multicolumn{1}{l|}{\cellcolor[HTML]{D9EAD3}\textbf{43.0}} &
  \multicolumn{1}{l|}{\cellcolor[HTML]{D9EAD3}51.0} &
  \cellcolor[HTML]{D9EAD3}6.0 &
  \multicolumn{1}{l|}{\cellcolor[HTML]{D9EAD3}{\underline{\textbf{70.0}}}} &
  \multicolumn{1}{l|}{\cellcolor[HTML]{D9EAD3}30.0} &
  \cellcolor[HTML]{D9EAD3}0.0 &
  \multicolumn{1}{l|}{\cellcolor[HTML]{D9EAD3}\textbf{36.0}} &
  \multicolumn{1}{l|}{\cellcolor[HTML]{D9EAD3}59.0} &
  \cellcolor[HTML]{D9EAD3}5.0 \\ \cline{2-17} 
 &
  \cellcolor[HTML]{C9DAF8}\textbf{monotonicity negatives} &
  \multicolumn{1}{l|}{\cellcolor[HTML]{C9DAF8}5.0} &
  \multicolumn{1}{l|}{\cellcolor[HTML]{C9DAF8}\textbf{94.0}} &
  \cellcolor[HTML]{C9DAF8}1.0 &
  \multicolumn{1}{l|}{\cellcolor[HTML]{C9DAF8}22.0} &
  \multicolumn{1}{l|}{\cellcolor[HTML]{C9DAF8}\textbf{28.0}} &
  \cellcolor[HTML]{C9DAF8}50.0 &
  \multicolumn{1}{l|}{\cellcolor[HTML]{C9DAF8}37.0} &
  \multicolumn{1}{l|}{\cellcolor[HTML]{C9DAF8}\textbf{42.0}} &
  \cellcolor[HTML]{C9DAF8}21.0 &
  \multicolumn{1}{l|}{\cellcolor[HTML]{C9DAF8}33.0} &
  \multicolumn{1}{l|}{\cellcolor[HTML]{C9DAF8}{\underline{\textbf{56.0}}}} &
  \cellcolor[HTML]{C9DAF8}11.0 &
  \multicolumn{1}{l|}{\cellcolor[HTML]{C9DAF8}46.0} &
  \multicolumn{1}{l|}{\cellcolor[HTML]{C9DAF8}\textbf{35.0}} &
  \cellcolor[HTML]{C9DAF8}19.0 \\ \cline{2-17} 
\multirow{-4}{*}{\textbf{Standalone}} &
  \cellcolor[HTML]{C9DAF8}\textbf{uncertainty adverbs} &
  \multicolumn{1}{l|}{\cellcolor[HTML]{C9DAF8}0.0} &
  \multicolumn{1}{l|}{\cellcolor[HTML]{C9DAF8}\textbf{100.0}} &
  \cellcolor[HTML]{C9DAF8}0.0 &
  \multicolumn{1}{l|}{\cellcolor[HTML]{C9DAF8}77.6} &
  \multicolumn{1}{l|}{\cellcolor[HTML]{C9DAF8}\textbf{4.7}} &
  \cellcolor[HTML]{C9DAF8}17.8 &
  \multicolumn{1}{l|}{\cellcolor[HTML]{C9DAF8}79.9} &
  \multicolumn{1}{l|}{\cellcolor[HTML]{C9DAF8}\textbf{9.1}} &
  \cellcolor[HTML]{C9DAF8}11.0 &
  \multicolumn{1}{l|}{\cellcolor[HTML]{C9DAF8}61.0} &
  \multicolumn{1}{l|}{\cellcolor[HTML]{C9DAF8}{\underline{\textbf{38.0}}}} &
  \cellcolor[HTML]{C9DAF8}1.0 &
  \multicolumn{1}{l|}{\cellcolor[HTML]{C9DAF8}89.9} &
  \multicolumn{1}{l|}{\cellcolor[HTML]{C9DAF8}\textbf{9.0}} &
  \cellcolor[HTML]{C9DAF8}1.1 \\ \hline
 &
  \cellcolor[HTML]{D9EAD3}\textbf{pronouns} &
  \multicolumn{1}{l|}{\cellcolor[HTML]{D9EAD3}\textbf{100.0}} &
  \multicolumn{1}{l|}{\cellcolor[HTML]{D9EAD3}0.0} &
  \cellcolor[HTML]{D9EAD3}0.0 &
  \multicolumn{1}{l|}{\cellcolor[HTML]{D9EAD3}{\color[HTML]{6AA84F} \textbf{65.4}}} &
  \multicolumn{1}{l|}{\cellcolor[HTML]{D9EAD3}{\color[HTML]{FF0000} \textit{27.8}}} &
  \cellcolor[HTML]{D9EAD3}{\color[HTML]{6AA84F} 6.8} &
  \multicolumn{1}{l|}{\cellcolor[HTML]{D9EAD3}{\color[HTML]{6AA84F} \textit{\textbf{72.3}}}} &
  \multicolumn{1}{l|}{\cellcolor[HTML]{D9EAD3}{\color[HTML]{FF0000} 23.3}} &
  \cellcolor[HTML]{D9EAD3}{\color[HTML]{FF0000} 4.3} &
  \multicolumn{1}{l|}{\cellcolor[HTML]{D9EAD3}{\underline{\textit{\textbf{100.0}}}}} &
  \multicolumn{1}{l|}{\cellcolor[HTML]{D9EAD3}0.0} &
  \cellcolor[HTML]{D9EAD3}0.0 &
  \multicolumn{1}{l|}{\cellcolor[HTML]{D9EAD3}{\color[HTML]{6AA84F} \textit{\textbf{32.0}}}} &
  \multicolumn{1}{l|}{\cellcolor[HTML]{D9EAD3}{\color[HTML]{FF0000} 58.1}} &
  \cellcolor[HTML]{D9EAD3}{\color[HTML]{6AA84F} 9.9} \\ \cline{2-17} 
 &
  \cellcolor[HTML]{D9EAD3}\textbf{monotonicity positives} &
  \multicolumn{1}{l|}{\cellcolor[HTML]{D9EAD3}{\color[HTML]{6AA84F} \textbf{97.0}}} &
  \multicolumn{1}{l|}{\cellcolor[HTML]{D9EAD3}{\color[HTML]{FF0000} 3.0}} &
  \cellcolor[HTML]{D9EAD3}0.0 &
  \multicolumn{1}{l|}{\cellcolor[HTML]{D9EAD3}{\color[HTML]{6AA84F} \textbf{38.0}}} &
  \multicolumn{1}{l|}{\cellcolor[HTML]{D9EAD3}{\color[HTML]{FF0000} 25.1}} &
  \cellcolor[HTML]{D9EAD3}{\color[HTML]{6AA84F} 36.8} &
  \multicolumn{1}{l|}{\cellcolor[HTML]{D9EAD3}{\color[HTML]{6AA84F} \textit{\textbf{55.7}}}} &
  \multicolumn{1}{l|}{\cellcolor[HTML]{D9EAD3}{\color[HTML]{FF0000} 37.2}} &
  \cellcolor[HTML]{D9EAD3}{\color[HTML]{6AA84F} 7.1} &
  \multicolumn{1}{l|}{\cellcolor[HTML]{D9EAD3}{\color[HTML]{6AA84F} {\underline{\textit{\textbf{71.0}}}}}} &
  \multicolumn{1}{l|}{\cellcolor[HTML]{D9EAD3}{\color[HTML]{FF0000} 29.0}} &
  \cellcolor[HTML]{D9EAD3}0.0 &
  \multicolumn{1}{l|}{\cellcolor[HTML]{D9EAD3}{\color[HTML]{6AA84F} \textit{\textbf{36.9}}}} &
  \multicolumn{1}{l|}{\cellcolor[HTML]{D9EAD3}{\color[HTML]{FF0000} 44.9}} &
  \cellcolor[HTML]{D9EAD3}{\color[HTML]{6AA84F} 18.3} \\ \cline{2-17} 
 &
  \cellcolor[HTML]{C9DAF8}\textbf{monotonicity negatives} &
  \multicolumn{1}{l|}{\cellcolor[HTML]{C9DAF8}5.0} &
  \multicolumn{1}{l|}{\cellcolor[HTML]{C9DAF8}{\color[HTML]{6AA84F} \textbf{95.0}}} &
  \cellcolor[HTML]{C9DAF8}{\color[HTML]{FF0000} 0.0} &
  \multicolumn{1}{l|}{\cellcolor[HTML]{C9DAF8}{\color[HTML]{6AA84F} 39.8}} &
  \multicolumn{1}{l|}{\cellcolor[HTML]{C9DAF8}{\color[HTML]{FF0000} \textbf{19.0}}} &
  \cellcolor[HTML]{C9DAF8}{\color[HTML]{FF0000} 41.2} &
  \multicolumn{1}{l|}{\cellcolor[HTML]{C9DAF8}{\color[HTML]{6AA84F} 52.8}} &
  \multicolumn{1}{l|}{\cellcolor[HTML]{C9DAF8}{\color[HTML]{FF0000} \textbf{29.4}}} &
  \cellcolor[HTML]{C9DAF8}{\color[HTML]{FF0000} 17.8} &
  \multicolumn{1}{l|}{\cellcolor[HTML]{C9DAF8}{\color[HTML]{6AA84F} 36.0}} &
  \multicolumn{1}{l|}{\cellcolor[HTML]{C9DAF8}{\color[HTML]{FF0000} {\underline{\textbf{53.0}}}}} &
  \cellcolor[HTML]{C9DAF8}11.0 &
  \multicolumn{1}{l|}{\cellcolor[HTML]{C9DAF8}{\color[HTML]{FF0000} 32.4}} &
  \multicolumn{1}{l|}{\cellcolor[HTML]{C9DAF8}{\color[HTML]{6AA84F} \textbf{38.5}}} &
  \cellcolor[HTML]{C9DAF8}{\color[HTML]{6AA84F} 29.0} \\ \cline{2-17} 
\multirow{-4}{*}{\textbf{\begin{tabular}[c]{@{}l@{}}Under presupposition\\ triggers\end{tabular}}} &
  \cellcolor[HTML]{C9DAF8}\textbf{uncertainty adverbs (sample)} &
  \multicolumn{1}{l|}{\cellcolor[HTML]{C9DAF8}{\color[HTML]{6AA84F} 1.0}} &
  \multicolumn{1}{l|}{\cellcolor[HTML]{C9DAF8}{\color[HTML]{FF0000} \textbf{99.0}}} &
  \cellcolor[HTML]{C9DAF8}0.0 &
  \multicolumn{1}{l|}{\cellcolor[HTML]{C9DAF8}{\color[HTML]{6AA84F} 87.8}} &
  \multicolumn{1}{l|}{\cellcolor[HTML]{C9DAF8}{\color[HTML]{FF0000} \textbf{2.7}}} &
  \cellcolor[HTML]{C9DAF8}{\color[HTML]{FF0000} 9.6} &
  \multicolumn{1}{l|}{\cellcolor[HTML]{C9DAF8}{\color[HTML]{6AA84F} 86.9}} &
  \multicolumn{1}{l|}{\cellcolor[HTML]{C9DAF8}{\color[HTML]{FF0000} \textbf{7.5}}} &
  \cellcolor[HTML]{C9DAF8}{\color[HTML]{FF0000} 5.6} &
  \multicolumn{1}{l|}{\cellcolor[HTML]{C9DAF8}{\color[HTML]{6AA84F} 70.0}} &
  \multicolumn{1}{l|}{\cellcolor[HTML]{C9DAF8}{\color[HTML]{FF0000} {\underline{\textbf{30.0}}}}} &
  \cellcolor[HTML]{C9DAF8}{\color[HTML]{FF0000} 0.0} &
  \multicolumn{1}{l|}{\cellcolor[HTML]{C9DAF8}{\color[HTML]{FF0000} 58.4}} &
  \multicolumn{1}{l|}{\cellcolor[HTML]{C9DAF8}{\color[HTML]{6AA84F} \textbf{29.4}}} &
  \cellcolor[HTML]{C9DAF8}{\color[HTML]{6AA84F} 12.3} \\ \hline
 &
  \cellcolor[HTML]{C9DAF8}\textbf{pronouns} &
  \multicolumn{1}{l|}{\cellcolor[HTML]{C9DAF8}{\color[HTML]{FF0000} 0.0}} &
  \multicolumn{1}{l|}{\cellcolor[HTML]{C9DAF8}{\color[HTML]{6AA84F} \textbf{100.0}}} &
  \cellcolor[HTML]{C9DAF8}0.0 &
  \multicolumn{1}{l|}{\cellcolor[HTML]{C9DAF8}{\color[HTML]{6AA84F} 51.0}} &
  \multicolumn{1}{l|}{\cellcolor[HTML]{C9DAF8}{\color[HTML]{FF0000} \textbf{41.0}}} &
  \cellcolor[HTML]{C9DAF8}{\color[HTML]{6AA84F} 8.0} &
  \multicolumn{1}{l|}{\cellcolor[HTML]{C9DAF8}{\color[HTML]{6AA84F} 56.7}} &
  \multicolumn{1}{l|}{\cellcolor[HTML]{C9DAF8}{\color[HTML]{6AA84F} \textbf{40.0}}} &
  \cellcolor[HTML]{C9DAF8}{\color[HTML]{FF0000} 3.3} &
  \multicolumn{1}{l|}{\cellcolor[HTML]{C9DAF8}{\color[HTML]{FF0000} 97.0}} &
  \multicolumn{1}{l|}{\cellcolor[HTML]{C9DAF8}{\color[HTML]{6AA84F} \textbf{3.0}}} &
  \cellcolor[HTML]{C9DAF8}0.0 &
  \multicolumn{1}{l|}{\cellcolor[HTML]{C9DAF8}{\color[HTML]{FF0000} 29.7}} &
  \multicolumn{1}{l|}{\cellcolor[HTML]{C9DAF8}{\color[HTML]{6AA84F} {\underline{\textbf{64.7}}}}} &
  \cellcolor[HTML]{C9DAF8}{\color[HTML]{FF0000} 5.5} \\ \cline{2-17} 
 &
  \cellcolor[HTML]{C9DAF8}\textbf{monotonicity positives} &
  \multicolumn{1}{l|}{\cellcolor[HTML]{C9DAF8}{\color[HTML]{FF0000} 1.0}} &
  \multicolumn{1}{l|}{\cellcolor[HTML]{C9DAF8}{\color[HTML]{6AA84F} \textbf{99.0}}} &
  \cellcolor[HTML]{C9DAF8}{\color[HTML]{6AA84F} 1.0} &
  \multicolumn{1}{l|}{\cellcolor[HTML]{C9DAF8}{\color[HTML]{6AA84F} 28.8}} &
  \multicolumn{1}{l|}{\cellcolor[HTML]{C9DAF8}{\color[HTML]{FF0000} \textbf{27.3}}} &
  \cellcolor[HTML]{C9DAF8}{\color[HTML]{6AA84F} 43.9} &
  \multicolumn{1}{l|}{\cellcolor[HTML]{C9DAF8}{\color[HTML]{6AA84F} 60.2}} &
  \multicolumn{1}{l|}{\cellcolor[HTML]{C9DAF8}{\color[HTML]{FF0000} \textbf{30.2}}} &
  \cellcolor[HTML]{C9DAF8}{\color[HTML]{6AA84F} 9.7} &
  \multicolumn{1}{l|}{\cellcolor[HTML]{C9DAF8}{\color[HTML]{FF0000} 44.0}} &
  \multicolumn{1}{l|}{\cellcolor[HTML]{C9DAF8}{\color[HTML]{6AA84F} {\underline{\textbf{56.0}}}}} &
  \cellcolor[HTML]{C9DAF8}0.0 &
  \multicolumn{1}{l|}{\cellcolor[HTML]{C9DAF8}{\color[HTML]{FF0000} 34.4}} &
  \multicolumn{1}{l|}{\cellcolor[HTML]{C9DAF8}{\color[HTML]{FF0000} \textbf{52.3}}} &
  \cellcolor[HTML]{C9DAF8}{\color[HTML]{6AA84F} 13.3} \\ \cline{2-17} 
\multirow{-3}{*}{\textbf{\begin{tabular}[c]{@{}l@{}}Under\\ non-factives\end{tabular}}} &
  \cellcolor[HTML]{C9DAF8}\textbf{monotonicity negatives} &
  \multicolumn{1}{l|}{\cellcolor[HTML]{C9DAF8}{\color[HTML]{FF0000} 0.0}} &
  \multicolumn{1}{l|}{\cellcolor[HTML]{C9DAF8}{\color[HTML]{6AA84F} \textbf{100.0}}} &
  \cellcolor[HTML]{C9DAF8}{\color[HTML]{FF0000} 0.0} &
  \multicolumn{1}{l|}{\cellcolor[HTML]{C9DAF8}{\color[HTML]{6AA84F} 30.3}} &
  \multicolumn{1}{l|}{\cellcolor[HTML]{C9DAF8}{\color[HTML]{FF0000} \textbf{17.4}}} &
  \cellcolor[HTML]{C9DAF8}{\color[HTML]{6AA84F} 52.3} &
  \multicolumn{1}{l|}{\cellcolor[HTML]{C9DAF8}{\color[HTML]{6AA84F} 52.9}} &
  \multicolumn{1}{l|}{\cellcolor[HTML]{C9DAF8}{\color[HTML]{FF0000} \textbf{22.3}}} &
  \cellcolor[HTML]{C9DAF8}{\color[HTML]{6AA84F} 24.7} &
  \multicolumn{1}{l|}{\cellcolor[HTML]{C9DAF8}{\color[HTML]{FF0000} 23.0}} &
  \multicolumn{1}{l|}{\cellcolor[HTML]{C9DAF8}{\color[HTML]{6AA84F} {\underline{\textbf{69.0}}}}} &
  \cellcolor[HTML]{C9DAF8}{\color[HTML]{FF0000} 8.0} &
  \multicolumn{1}{l|}{\cellcolor[HTML]{C9DAF8}{\color[HTML]{FF0000} 29.3}} &
  \multicolumn{1}{l|}{\cellcolor[HTML]{C9DAF8}{\color[HTML]{6AA84F} \textbf{42.0}}} &
  \cellcolor[HTML]{C9DAF8}{\color[HTML]{6AA84F} 28.7} \\ \hline
\end{tabular}
}

\caption{Experiment results across all models. Background indicates the expected label (green - "entailment", blue - "neutral"). The green and red font colors indicate increase or decrease compared to standalone premises. The numbers in \textbf{bold} also indicate \textit{accuracy} (e.g. for uncertainty adverbs accuracy coincides with the percentage of neutrals). The \underline{\textbf{underlined bold}} numbers indicate \textit{the best accuracy for a specific inference type} across all models (human baseline excluded). \textbf{e} - entailment (\%); \textbf{n} - neutral (\%); \textbf{c} - contradiction (\%). Model versions shown: GPT-3.5 -- text-davinci-003; ChatGPT -- gpt-3.5-turbo-0301; GPT-4 - gpt-4-0613. Result for additional GPT models (gpt-3.5-turbo-0613 and gpt-4-0314) available in Appendix \ref{sec:acc}.}
\label{tab:all1}
\end{table*}

\section{Prompt Variations}
\label{additional_exp}
We also experiment with prompt variations.

\paragraph{Prompt paraphrasing.} \label{consistency} We ask ChatGPT (gpt-3.5-turbo-0301) to rephrase our prompt template, obtaining two templates that we verify to be semantically equivalent w.r.t.\ the task (details in Appendix \ref{sec:consistency}).
While accuracies vary between prompts, the scores on all tasks remain low.

\paragraph{Chain-of-Thought Prompting (CoT).} Using gpt-3.5-turbo-0301,
we investigate CoT prompting \citep{kojima2023large}, and find that 
it \emph{reverses} the LLM's trend to predict more entailment for embedded premises
(see Section \ref{experiments}): now the LLM predicts \textit{less} entailment
under presupposition triggers or non-factives, while the number of neutral predictions grows - again, regardless of the correct label, i.e. the reversed trend is equally inaccurate. CoT prompting improves the accuracy, but in a one-sided way: scoring higher on ``neutral'' test sets, but lower on almost all the ``entailment'' ones.

\paragraph{Analysis of CoT Results.} The CoT technique allows us to explore the model's ``reasoning''. We manually evaluate a subset of the CoT explanations. In half of the cases (50.9\%) both the final decision and the CoT explanation were wrong. In 23.6\% a correct explanation was followed by a correct decision; in
23.6\%
a wrong explanation was followed by a correct decision. In 1.86\% of the cases a correct
explanation was followed by a wrong
decision.
81\% of the cases reflected a correct understanding of the task expressed in the prompt. 
In half of the cases (49.1\%) the CoT mentioned the undelying linguistic phenomena explicitly, but only in half of those (23.6\% of the total) reflected their correct understanding and only 14.5\% of the time used them as a basis for the final prediction.

The details of the CoT experiments and the manual analysis are available in Appendix \ref{sec:cot}.

\section{Relation to previous findings}

A vast literature evaluated linguistic abilities of earlier generations of pretrained models\footnote{There are comprehensive overviews of such literature, for example, \citet{rogers2020primer,chang_etal}.}, including some phenomena we consider here: \citet{jiang-de-marneffe-2019-evaluating,de_Marneffe_Simons_Tonhauser_2019, ross-pavlick-2019-well,mccoy-etal-2019-right,yanaka2021exploring} (\textbf{embedded clauses}); \citet{jeretic-etal-2020-natural, parrish-etal-2021-nope, kabbara-cheung-2022-investigating} (\textbf{presuppositions and implicatures});\citet{yanaka-etal-2019-neural, yanaka-etal-2019-help, goodwin2020probing, yanaka-etal-2020-neural,jumelet-etal-2021-language, geiger2020neural} (\textbf{monotonicity entailment}); \citet{Ettinger, guerin-chemla-2023-bird} (\textbf{hypernymy}). Some works focused on other phenomena: \textbf{negation, synonyms, antonyms, tenses, temporal adverbials, active vs. passive voice, symmetric and asymmetric relations, comparatives, superlatives} etc. \citeg{kassner-schutze-2020-negated,hartmann-etal-2021-multilingual, hossain-etal-2020-analysis, hossain-etal-2022-analysis,Ettinger,truong-etal-2022-improving, jang-etal-2022-beyond, ravichander2022condaqa,ribeiro-etal-2020-beyond}. Many of these studies make use of the NLI task to evaluate LMs' basic linguistic abilities \citeg{mccoy-etal-2019-right, jiang-de-marneffe-2019-evaluating, yanaka2021exploring, kabbara-cheung-2022-investigating,jeretic-etal-2020-natural, parrish-etal-2021-nope, kabbara-cheung-2022-investigating,yanaka-etal-2019-neural, yanaka-etal-2019-help}. We draw inspiration from this line of work.

In the remainder of the section, we focus on work that evaluates SOTA LLMs (starting with GPT-3).

\citet{truong-etal-2023-language} seek to determine if \textbf{negation} remains a challenge in the age of LLMs.\footnote{In fact, their work covers a range of negation-related phenomena, such as synonym/antonym lexical relations, monotonicity entailment and hypernymy under negation.} They assess GPTneo, GPT-3, and InstructGPT across various negation benchmarks with different model sizes and prompts, and discover a number of important limitations, including insensitivity to negation, an inability to capture synonym/antonym lexical relations and to reason under negation.
\citet{ye-etal-2023-assessing} find that LLMs\footnote{They test 14 LLMs including GPT-4 and 3.5 and four LLaMA variants.} with CoT-style prompts struggle with \textbf{lexical negation} (words with \textbf{negative suffixes}: \textit{implausible}, \textit{unreasonable} etc.).
\citet{Lorge2023NotWV} explore the capability of LMs (including text-davinci-002) to handle \textbf{scalar adverbs} (specifically, \textbf{adverbs of frequency, modality and degree}).
Despite some success they notice many shortcomings: “weak differentiation amongst the semantic classes of adverbs, poor ability to discriminate scalar adverbs from
negations”, “strong effects of adverb frequencies and lack of generalisation across two logically equivalent entailment constructions”.
\citet{cong-2022-psycholinguistic}
probes GPT-3 for understanding of scalar \textbf{implicatures} and \textbf{presuppositions} and find that its performance is ``mostly at chance''.
\citet{stengel-eskin-van-durme-2022-curious} investigate how GPT3-davinci, GPT-Neo, Jurassic, T5 for QA, T0pp handle\textbf{ subject-control sentences}.
\citet{davis-2022-incremental} analyzes how autoregressive models with GPT-like architectures, such as GPT-2 XL, GPT-Neo, GPT-J, GPT-3 (text-davinci-002) handle \textbf{binding} and conclude that they
consider antecedents which humans reject as structurally ungrammatical.
Focusing on “language illusions” , \citet{Zhang2023CanLM} find that LMs including text-davinci-003
“have some
capacity to process \textbf{comparative structures}”, but struggle with \textbf{multiple negation} and \textbf{NPI licensors}.

This overview suggests that 
for newer LLMs (GPT-3 and later) there is much less research focusing on fundamental linguistic abilities. We add to this line of work, focusing on very simple inference types that to the best of our knowledge were not previously evaluated (grammatically-specified entailments, uncetainty adverbs) as well as evaluating stability under embedding contexts (presupposition-triggers and non-factives).

At the same time, the existing research shows that current pre-training objectives do not necessarily capture the basic linguistic phenomena, thus ``new training paradigms are essential to achieve better linguistic competency'' \citep{truong-etal-2023-language}.

\section{Conclusion}
We tested LLMs with several simple linguistic inferences that humans find trivial, and further experimented by embedding the premises within (simple) linguistic contexts that should either preserve or cancel the inference. The models performed poorly in all cases. We conclude that LLMs do not learn entailment semantics ``naturally''.
 The persistence of the issue across prompts, models and setups shows that these limitations are robust and this topic merits further systematic investigations. 
 While the results are tested for several specific models, we do expect them to hold more generally. We believe the ability of LLMs to correctly understand simple textual inferences humans find natural is important. We provide a benchmark and methodology that can be used to track the improvement of future models.

\section{Limitations}

As a consequence of LLMs' sensitivity to prompt phrasings, there may exist prompts that can potentially modify the reported results.
At the same time we agree with \citet{jang2023consistency} who point out that ``improvements with prompt design can be considered another violation of semantic consistency, because the prompts will deliver identical semantic meaning, i.e., task description''.

We showed that embedding the premises under presupposition triggers or non-factives affects the models' predictions exhibiting certain patterns. However, our treatment of non-factives is non-exhaustive: for example, we did not try adding negation to the non-factives or using non-factive predicates denoting a high level of uncertainty (\textit{I doubt, I'm skeptical}) or containing negative prefixes (\textit{I an uncertain, I disbelieve}). It's possible that implicit or explicit negation in the embedding predicates may change the LLMs' behavior. Also, while considering presupposition triggers and non-factives, we did not consider other types of clause-embedding predicates (e.g., implicative verbs).

Finally, since LLMs undergo continuous updates, the test results presented here may vary over time, and the closed, black-box nature of the models' training details and training data limits the scope of the conclusions that can be reached. We cannot, for example, know for certain if the low performance on these inference type stems already from the base model and its inability to learn these type of inferences from raw text, or was added during instruct tuning.  
That said, our data and methodology for benchmarking these capabilities is model-agnostic and remains intact.

\section*{Ethics Statement}
\paragraph{Annotator Qualification and Compensation for Annotation} To collect annotations for our human baseline (see \S\ref{human}) we used Amazon Mechanical Turk (AMT). We employed 5 annotators who showed themselves reliable in handling annotation tasks assigned by us in the past. All annotators had the following qualifications: (1) over 500 completed HITs; (2) 99\% approval rate or higher; (3) native English speakers. We paid \$0.10 per example. Labeling each example normally (excluding outliers) took about 21.7 seconds which resulted in a payment of about \$16.5 per hour. Upon completing the task the annotators were given bonuses.
\paragraph{Annotation Data Collection and Usage Policy} Annotators were notified that their annotations are meant for research purposes in the field of Natural Language Processing. The task and collected annotations were objective and included no personal information.

\bibliography{anthology,custom}

\clearpage
\appendix

\section{Human annotation details}
\label{sec:h_baseline}
In order to construct a human baseline, we sampled 1100 examples (100 per inference type) and had 3 human annotators label each example, asking if based on Text 1, Text 2 is true, false or neutral.
We then use majority vote to determine the final human label for each example. 

For human annotation, we used the Amazon Mechanical Turk platform. We did not use any qualification tests and filters for annotator selection, but instead employed 5 annotators who showed themselves reliable in handling annotation tasks assigned by us in the past. We paid \$0.10 per example. Labeling each example normally (excluding outliers) took about 21.7 seconds which resulted in a payment of about \$16.5 per hour.

We obtained an overall accuracy of 97.55\%. In 91\% of the cases all three annotators confirmed the expected labels unanimously.
The instructions received by the annotators are shown in Figure \ref{fig:instr}.
\begin{figure*}[t!]
\includegraphics[width=\textwidth]{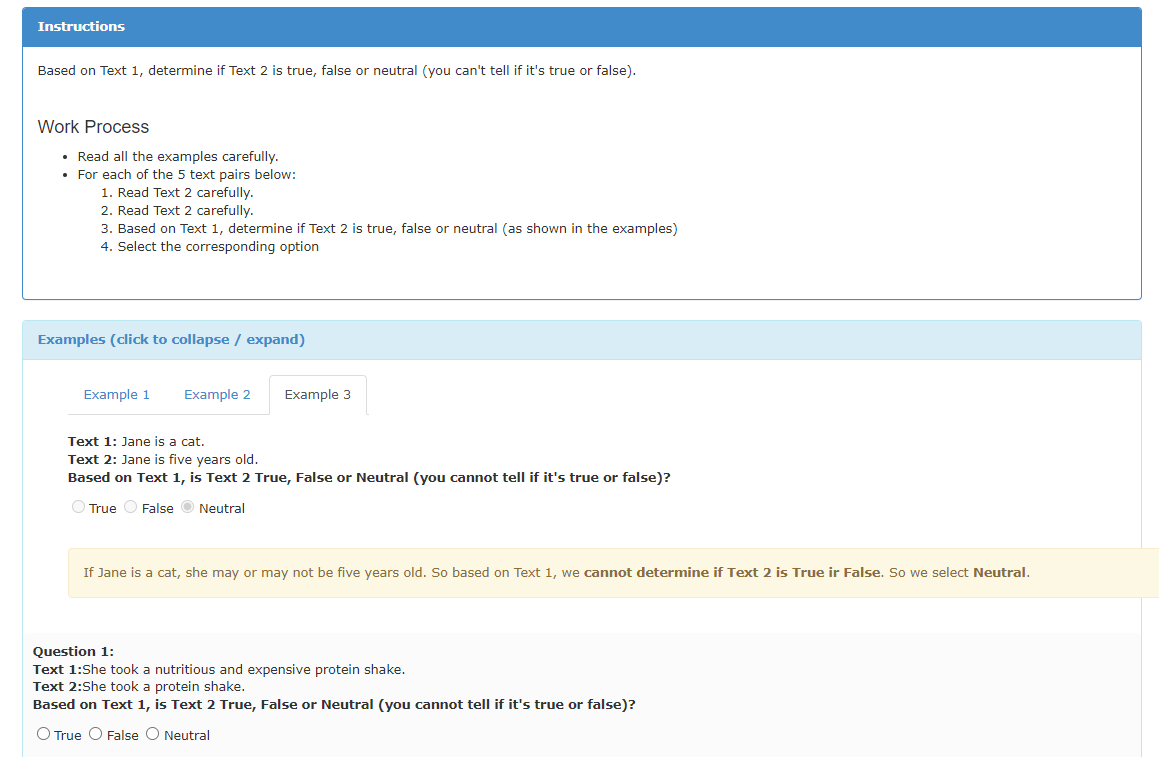}
\caption{Instructions for human annotation. Note that in order to not implicitly train annotators towards the linguistic inferences we consider in this work, we provided examples that demonstrate the meaning of the neutral/entailing/contradiction labels, but on other inference types, not covered in this work.}
\label{fig:instr}
\end{figure*}

\section{Main entailment experiments prompts}
\label{sec:orig_prompt}
The prompt below is used throughout the experiments described in Section \ref{experiments} for all the OpenAI models (see Section \ref{experiments}).
\begin{quote}
\textit{You are given a pair of texts. Say about this pair: given Text 1, is Text 2 true, false or neutral (you can't tell if it's true or false)? Reply in one word.}

 \textit{Text 1: ``{text1}''}
 
 \textit{Text 2: ``{text2}'' }
 
\end{quote} 

For LLaMA 2 we use the following prompt:

\begin{quote}
\textit{You have two texts, and your task is to determine the truthfulness of Text 2 based on Text 1. Provide a one-word response indicating whether Text 2 is true, false, or neutral (indeterminable). Here are the texts:}

 \textit{Text 1: ``{text1}''}
 
 \textit{Text 2: ``{text2}''}

\end{quote} 

The model outputs one of three possible labels: \textit{“true”} (corresponding to \textit{“entailment”}), \textit{“false”} (corresponding to \textit{“contradiction”}) or \textit{“neutral”}.\footnote{In the rare cases when the model outputs a different label, we normalize it to one of the three expected forms. E.g. ``truthful'' is normalized to ``true''.}

\section{Complete Results Table}\label{sec:acc}
All the results from all LLMs including both ChatGPT and GPT-4 versions are shown in Table \ref{tab:all}.

\begin{table*}[t!]

\resizebox{\textwidth}{!}{%

\begin{tabular}{|l|l|b{0.75cm}b{0.75cm}b{0.75cm}|b{0.75cm}b{0.75cm}b{0.75cm}|b{0.75cm}b{0.75cm}b{0.75cm}|b{0.75cm}b{0.75cm}b{0.75cm}|b{0.75cm}b{0.75cm}b{0.75cm}|b{0.75cm}b{0.75cm}b{0.75cm}|b{0.75cm}b{0.75cm}b{0.75cm}|}
\hline
 &
   &
  \multicolumn{3}{l|}{\textbf{human}} &
  \multicolumn{3}{l|}{\textbf{text-davinci-003}} &
  \multicolumn{3}{l|}{\textbf{gpt-3.5-turbo-0301}} &
  \multicolumn{3}{l|}{\textbf{gpt-3.5-turbo-0613}} &
  \multicolumn{3}{l|}{\textbf{gpt-4-0314}} &
  \multicolumn{3}{l|}{\textbf{gpt-4-0613}} &
  \multicolumn{3}{l|}{\textbf{LLaMA 2 70b}} \\ \hline
 &
   &
  \multicolumn{1}{b{0.75cm}|}{\textbf{e}} &
  \multicolumn{1}{b{0.75cm}|}{\textbf{n}} &
  \textbf{c} &
  \multicolumn{1}{b{0.75cm}|}{\textbf{e}} &
  \multicolumn{1}{b{0.75cm}|}{\textbf{n}} &
  \textbf{c} &
  \multicolumn{1}{b{0.75cm}|}{\textbf{e}} &
  \multicolumn{1}{b{0.75cm}|}{\textbf{n}} &
  \textbf{c} &
  \multicolumn{1}{b{0.75cm}|}{\textbf{e}} &
  \multicolumn{1}{b{0.75cm}|}{\textbf{n}} &
  \textbf{c} &
  \multicolumn{1}{b{0.75cm}|}{\textbf{e}} &
  \multicolumn{1}{b{0.75cm}|}{\textbf{n}} &
  \textbf{c} &
  \multicolumn{1}{b{0.75cm}|}{\textbf{e}} &
  \multicolumn{1}{b{0.75cm}|}{\textbf{n}} &
  \textbf{c} &
  \multicolumn{1}{b{0.75cm}|}{\textbf{e}} &
  \multicolumn{1}{b{0.75cm}|}{\textbf{n}} &
  \textbf{c} \\ \hline
 &
  \cellcolor[HTML]{D9EAD3}\textbf{P} &
  \multicolumn{1}{b{0.75cm}|}{\cellcolor[HTML]{D9EAD3}\textbf{100}} &
  \multicolumn{1}{b{0.75cm}|}{\cellcolor[HTML]{D9EAD3}0} &
  \cellcolor[HTML]{D9EAD3}0 &
  \multicolumn{1}{b{0.75cm}|}{\cellcolor[HTML]{D9EAD3}\textbf{39}} &
  \multicolumn{1}{b{0.75cm}|}{\cellcolor[HTML]{D9EAD3}56} &
  \cellcolor[HTML]{D9EAD3}5 &
  \multicolumn{1}{b{0.75cm}|}{\cellcolor[HTML]{D9EAD3}\textbf{53}} &
  \multicolumn{1}{b{0.75cm}|}{\cellcolor[HTML]{D9EAD3}34} &
  \cellcolor[HTML]{D9EAD3}13 &
  \multicolumn{1}{b{0.75cm}|}{\cellcolor[HTML]{D9EAD3}\textbf{9}} &
  \multicolumn{1}{b{0.75cm}|}{\cellcolor[HTML]{D9EAD3}53} &
  \cellcolor[HTML]{D9EAD3}38 &
  \multicolumn{1}{b{0.75cm}|}{\cellcolor[HTML]{D9EAD3}\textbf{84}} &
  \multicolumn{1}{b{0.75cm}|}{\cellcolor[HTML]{D9EAD3}16} &
  \cellcolor[HTML]{D9EAD3}0 &
  \multicolumn{1}{b{0.75cm}|}{\cellcolor[HTML]{D9EAD3}\underline{\textbf{100}}} &
  \multicolumn{1}{b{0.75cm}|}{\cellcolor[HTML]{D9EAD3}0} &
  \cellcolor[HTML]{D9EAD3}0 &
  \multicolumn{1}{b{0.75cm}|}{\cellcolor[HTML]{D9EAD3}\textbf{31}} &
  \multicolumn{1}{b{0.75cm}|}{\cellcolor[HTML]{D9EAD3}61} &
  \cellcolor[HTML]{D9EAD3}8 \\ \cline{2-23} 
 &
  \cellcolor[HTML]{D9EAD3}\textbf{MP} &
  \multicolumn{1}{b{0.75cm}|}{\cellcolor[HTML]{D9EAD3}\textbf{89}} &
  \multicolumn{1}{b{0.75cm}|}{\cellcolor[HTML]{D9EAD3}11} &
  \cellcolor[HTML]{D9EAD3}0 &
  \multicolumn{1}{b{0.75cm}|}{\cellcolor[HTML]{D9EAD3}\textbf{25}} &
  \multicolumn{1}{b{0.75cm}|}{\cellcolor[HTML]{D9EAD3}46} &
  \cellcolor[HTML]{D9EAD3}29 &
  \multicolumn{1}{b{0.75cm}|}{\cellcolor[HTML]{D9EAD3}\textbf{43}} &
  \multicolumn{1}{b{0.75cm}|}{\cellcolor[HTML]{D9EAD3}51} &
  \cellcolor[HTML]{D9EAD3}6 &
  \multicolumn{1}{b{0.75cm}|}{\cellcolor[HTML]{D9EAD3}\textbf{21}} &
  \multicolumn{1}{b{0.75cm}|}{\cellcolor[HTML]{D9EAD3}35} &
  \cellcolor[HTML]{D9EAD3}44 &
  \multicolumn{1}{b{0.75cm}|}{\cellcolor[HTML]{D9EAD3}\textbf{39}} &
  \multicolumn{1}{b{0.75cm}|}{\cellcolor[HTML]{D9EAD3}61} &
  \cellcolor[HTML]{D9EAD3}0 &
  \multicolumn{1}{b{0.75cm}|}{\cellcolor[HTML]{D9EAD3}\underline{\textbf{70}}} &
  \multicolumn{1}{b{0.75cm}|}{\cellcolor[HTML]{D9EAD3}30} &
  \cellcolor[HTML]{D9EAD3}0 &
  \multicolumn{1}{b{0.75cm}|}{\cellcolor[HTML]{D9EAD3}\textbf{36}} &
  \multicolumn{1}{b{0.75cm}|}{\cellcolor[HTML]{D9EAD3}59} &
  \cellcolor[HTML]{D9EAD3}5 \\ \cline{2-23} 
 &
  \cellcolor[HTML]{C9DAF8}\textbf{MN} &
  \multicolumn{1}{b{0.75cm}|}{\cellcolor[HTML]{C9DAF8}5} &
  \multicolumn{1}{b{0.75cm}|}{\cellcolor[HTML]{C9DAF8}\textbf{94}} &
  \cellcolor[HTML]{C9DAF8}1 &
  \multicolumn{1}{b{0.75cm}|}{\cellcolor[HTML]{C9DAF8}22} &
  \multicolumn{1}{b{0.75cm}|}{\cellcolor[HTML]{C9DAF8}\textbf{28}} &
  \cellcolor[HTML]{C9DAF8}50 &
  \multicolumn{1}{b{0.75cm}|}{\cellcolor[HTML]{C9DAF8}37} &
  \multicolumn{1}{b{0.75cm}|}{\cellcolor[HTML]{C9DAF8}\textbf{42}} &
  \cellcolor[HTML]{C9DAF8}21 &
  \multicolumn{1}{b{0.75cm}|}{\cellcolor[HTML]{C9DAF8}12} &
  \multicolumn{1}{b{0.75cm}|}{\cellcolor[HTML]{C9DAF8}\textbf{37}} &
  \cellcolor[HTML]{C9DAF8}51 &
  \multicolumn{1}{b{0.75cm}|}{\cellcolor[HTML]{C9DAF8}6} &
  \multicolumn{1}{b{0.75cm}|}{\cellcolor[HTML]{C9DAF8}\underline{\textbf{88}}} &
  \cellcolor[HTML]{C9DAF8}6 &
  \multicolumn{1}{b{0.75cm}|}{\cellcolor[HTML]{C9DAF8}33} &
  \multicolumn{1}{b{0.75cm}|}{\cellcolor[HTML]{C9DAF8}\textbf{56}} &
  \cellcolor[HTML]{C9DAF8}11 &
  \multicolumn{1}{b{0.75cm}|}{\cellcolor[HTML]{C9DAF8}46} &
  \multicolumn{1}{b{0.75cm}|}{\cellcolor[HTML]{C9DAF8}\textbf{35}} &
  \cellcolor[HTML]{C9DAF8}19 \\ \cline{2-23} 
\multirow{-4}{*}{\textbf{SA}} &
  \cellcolor[HTML]{C9DAF8}\textbf{UA} &
  \multicolumn{1}{b{0.75cm}|}{\cellcolor[HTML]{C9DAF8}0} &
  \multicolumn{1}{b{0.75cm}|}{\cellcolor[HTML]{C9DAF8}\textbf{100}} &
  \cellcolor[HTML]{C9DAF8}0 &
  \multicolumn{1}{b{0.75cm}|}{\cellcolor[HTML]{C9DAF8}77.56} &
  \multicolumn{1}{b{0.75cm}|}{\cellcolor[HTML]{C9DAF8}\textbf{4.67}} &
  \cellcolor[HTML]{C9DAF8}17.78 &
  \multicolumn{1}{b{0.75cm}|}{\cellcolor[HTML]{C9DAF8}79.89} &
  \multicolumn{1}{b{0.75cm}|}{\cellcolor[HTML]{C9DAF8}\textbf{9.11}} &
  \cellcolor[HTML]{C9DAF8}11.00 &
  \multicolumn{1}{b{0.75cm}|}{\cellcolor[HTML]{C9DAF8}34} &
  \multicolumn{1}{b{0.75cm}|}{\cellcolor[HTML]{C9DAF8}\textbf{19}} &
  \cellcolor[HTML]{C9DAF8}47 &
  \multicolumn{1}{b{0.75cm}|}{\cellcolor[HTML]{C9DAF8}30} &
  \multicolumn{1}{b{0.75cm}|}{\cellcolor[HTML]{C9DAF8}\underline{\textbf{70}}} &
  \cellcolor[HTML]{C9DAF8}0 &
  \multicolumn{1}{b{0.75cm}|}{\cellcolor[HTML]{C9DAF8}61} &
  \multicolumn{1}{b{0.75cm}|}{\cellcolor[HTML]{C9DAF8}\textbf{38}} &
  \cellcolor[HTML]{C9DAF8}1 &
  \multicolumn{1}{b{0.75cm}|}{\cellcolor[HTML]{C9DAF8}89.89} &
  \multicolumn{1}{b{0.75cm}|}{\cellcolor[HTML]{C9DAF8}\textbf{9}} &
  \cellcolor[HTML]{C9DAF8}1.11 \\ \hline
 &
  \cellcolor[HTML]{D9EAD3}\textbf{P} &
  \multicolumn{1}{b{0.75cm}|}{\cellcolor[HTML]{D9EAD3}\textbf{100}} &
  \multicolumn{1}{b{0.75cm}|}{\cellcolor[HTML]{D9EAD3}0} &
  \cellcolor[HTML]{D9EAD3}0 &
  \multicolumn{1}{b{0.75cm}|}{\cellcolor[HTML]{D9EAD3}{\color[HTML]{6AA84F} \textbf{65.35}}} &
  \multicolumn{1}{b{0.75cm}|}{\cellcolor[HTML]{D9EAD3}{\color[HTML]{FF0000} 27.83}} &
  \cellcolor[HTML]{D9EAD3}{\color[HTML]{6AA84F} 6.82} &
  \multicolumn{1}{b{0.75cm}|}{\cellcolor[HTML]{D9EAD3}{\color[HTML]{6AA84F} \textbf{72.30}}} &
  \multicolumn{1}{b{0.75cm}|}{\cellcolor[HTML]{D9EAD3}{\color[HTML]{FF0000} 23.35}} &
  \cellcolor[HTML]{D9EAD3}{\color[HTML]{FF0000} 4.35} &
  \multicolumn{1}{b{0.75cm}|}{\cellcolor[HTML]{D9EAD3}{\color[HTML]{FF0000} \textbf{6}}} &
  \multicolumn{1}{b{0.75cm}|}{\cellcolor[HTML]{D9EAD3}{\color[HTML]{FF0000} 42}} &
  \cellcolor[HTML]{D9EAD3}{\color[HTML]{6AA84F} 52} &
  \multicolumn{1}{b{0.75cm}|}{\cellcolor[HTML]{D9EAD3}{\color[HTML]{6AA84F} \textbf{90}}} &
  \multicolumn{1}{b{0.75cm}|}{\cellcolor[HTML]{D9EAD3}{\color[HTML]{FF0000} 10}} &
  \cellcolor[HTML]{D9EAD3}0 &
  \multicolumn{1}{b{0.75cm}|}{\cellcolor[HTML]{D9EAD3}\underline{\textbf{100}}} &
  \multicolumn{1}{b{0.75cm}|}{\cellcolor[HTML]{D9EAD3}0} &
  \cellcolor[HTML]{D9EAD3}0 &
  \multicolumn{1}{b{0.75cm}|}{\cellcolor[HTML]{D9EAD3}{\color[HTML]{6AA84F} \textbf{32.00}}} &
  \multicolumn{1}{b{0.75cm}|}{\cellcolor[HTML]{D9EAD3}{\color[HTML]{FF0000} 58.13}} &
  \cellcolor[HTML]{D9EAD3}{\color[HTML]{6AA84F} 9.87} \\ \cline{2-23} 
 &
  \cellcolor[HTML]{D9EAD3}\textbf{MP} &
  \multicolumn{1}{b{0.75cm}|}{\cellcolor[HTML]{D9EAD3}{\color[HTML]{6AA84F} \textbf{97}}} &
  \multicolumn{1}{b{0.75cm}|}{\cellcolor[HTML]{D9EAD3}{\color[HTML]{FF0000} 3}} &
  \cellcolor[HTML]{D9EAD3}0 &
  \multicolumn{1}{b{0.75cm}|}{\cellcolor[HTML]{D9EAD3}{\color[HTML]{6AA84F} \textbf{38.04}}} &
  \multicolumn{1}{b{0.75cm}|}{\cellcolor[HTML]{D9EAD3}{\color[HTML]{FF0000} 25.13}} &
  \cellcolor[HTML]{D9EAD3}{\color[HTML]{6AA84F} 36.83} &
  \multicolumn{1}{b{0.75cm}|}{\cellcolor[HTML]{D9EAD3}{\color[HTML]{6AA84F} \textbf{55.65}}} &
  \multicolumn{1}{b{0.75cm}|}{\cellcolor[HTML]{D9EAD3}{\color[HTML]{FF0000} 37.22}} &
  \cellcolor[HTML]{D9EAD3}{\color[HTML]{6AA84F} 7.13} &
  \multicolumn{1}{b{0.75cm}|}{\cellcolor[HTML]{D9EAD3}{\color[HTML]{FF0000} \textbf{17}}} &
  \multicolumn{1}{b{0.75cm}|}{\cellcolor[HTML]{D9EAD3}{\color[HTML]{6AA84F} 40}} &
  \cellcolor[HTML]{D9EAD3}{\color[HTML]{FF0000} 43} &
  \multicolumn{1}{b{0.75cm}|}{\cellcolor[HTML]{D9EAD3}\textbf{39}} &
  \multicolumn{1}{b{0.75cm}|}{\cellcolor[HTML]{D9EAD3}61} &
  \cellcolor[HTML]{D9EAD3}0 &
  \multicolumn{1}{b{0.75cm}|}{\cellcolor[HTML]{D9EAD3}{\color[HTML]{6AA84F} \underline{\textbf{71}}}} &
  \multicolumn{1}{b{0.75cm}|}{\cellcolor[HTML]{D9EAD3}{\color[HTML]{FF0000} 29}} &
  \cellcolor[HTML]{D9EAD3}0 &
  \multicolumn{1}{b{0.75cm}|}{\cellcolor[HTML]{D9EAD3}{\color[HTML]{6AA84F} \textbf{36.87}}} &
  \multicolumn{1}{b{0.75cm}|}{\cellcolor[HTML]{D9EAD3}{\color[HTML]{FF0000} 44.87}} &
  \cellcolor[HTML]{D9EAD3}{\color[HTML]{6AA84F} 18.26} \\ \cline{2-23} 
 &
  \cellcolor[HTML]{C9DAF8}\textbf{MN} &
  \multicolumn{1}{b{0.75cm}|}{\cellcolor[HTML]{C9DAF8}5} &
  \multicolumn{1}{b{0.75cm}|}{\cellcolor[HTML]{C9DAF8}{\color[HTML]{6AA84F} \textbf{95}}} &
  \cellcolor[HTML]{C9DAF8}{\color[HTML]{FF0000} 0} &
  \multicolumn{1}{b{0.75cm}|}{\cellcolor[HTML]{C9DAF8}{\color[HTML]{6AA84F} 39.78}} &
  \multicolumn{1}{b{0.75cm}|}{\cellcolor[HTML]{C9DAF8}{\color[HTML]{FF0000} \textbf{19.04}}} &
  \cellcolor[HTML]{C9DAF8}{\color[HTML]{FF0000} 41.18} &
  \multicolumn{1}{b{0.75cm}|}{\cellcolor[HTML]{C9DAF8}{\color[HTML]{6AA84F} 52.83}} &
  \multicolumn{1}{b{0.75cm}|}{\cellcolor[HTML]{C9DAF8}{\color[HTML]{FF0000} \textbf{29.39}}} &
  \cellcolor[HTML]{C9DAF8}{\color[HTML]{FF0000} 17.78} &
  \multicolumn{1}{b{0.75cm}|}{\cellcolor[HTML]{C9DAF8}{\color[HTML]{6AA84F} 21}} &
  \multicolumn{1}{b{0.75cm}|}{\cellcolor[HTML]{C9DAF8}{\color[HTML]{FF0000} \textbf{24}}} &
  \cellcolor[HTML]{C9DAF8}{\color[HTML]{6AA84F} 55} &
  \multicolumn{1}{b{0.75cm}|}{\cellcolor[HTML]{C9DAF8}{\color[HTML]{6AA84F} 9}} &
  \multicolumn{1}{b{0.75cm}|}{\cellcolor[HTML]{C9DAF8}{\color[HTML]{FF0000} \underline{\textbf{87}}}} &
  \cellcolor[HTML]{C9DAF8}{\color[HTML]{FF0000} 4} &
  \multicolumn{1}{b{0.75cm}|}{\cellcolor[HTML]{C9DAF8}{\color[HTML]{6AA84F} 36}} &
  \multicolumn{1}{b{0.75cm}|}{\cellcolor[HTML]{C9DAF8}{\color[HTML]{FF0000} \textbf{53}}} &
  \cellcolor[HTML]{C9DAF8}11 &
  \multicolumn{1}{b{0.75cm}|}{\cellcolor[HTML]{C9DAF8}{\color[HTML]{FF0000} 32.43}} &
  \multicolumn{1}{b{0.75cm}|}{\cellcolor[HTML]{C9DAF8}{\color[HTML]{6AA84F} \textbf{38.52}}} &
  \cellcolor[HTML]{C9DAF8}{\color[HTML]{6AA84F} 29.05} \\ \cline{2-23} 
\multirow{-4}{*}{\textbf{PT}} &
  \cellcolor[HTML]{C9DAF8}\textbf{UA} &
  \multicolumn{1}{b{0.75cm}|}{\cellcolor[HTML]{C9DAF8}{\color[HTML]{6AA84F} 1}} &
  \multicolumn{1}{b{0.75cm}|}{\cellcolor[HTML]{C9DAF8}{\color[HTML]{FF0000} \textbf{99}}} &
  \cellcolor[HTML]{C9DAF8}0 &
  \multicolumn{1}{b{0.75cm}|}{\cellcolor[HTML]{C9DAF8}{\color[HTML]{6AA84F} 87.78}} &
  \multicolumn{1}{b{0.75cm}|}{\cellcolor[HTML]{C9DAF8}{\color[HTML]{FF0000} \textbf{2.65}}} &
  \cellcolor[HTML]{C9DAF8}{\color[HTML]{FF0000} 9.57} &
  \multicolumn{1}{b{0.75cm}|}{\cellcolor[HTML]{C9DAF8}{\color[HTML]{6AA84F} 86.91}} &
  \multicolumn{1}{b{0.75cm}|}{\cellcolor[HTML]{C9DAF8}{\color[HTML]{FF0000} \textbf{7.52}}} &
  \cellcolor[HTML]{C9DAF8}{\color[HTML]{FF0000} 5.57} &
  \multicolumn{1}{b{0.75cm}|}{\cellcolor[HTML]{C9DAF8}{\color[HTML]{FF0000} 28}} &
  \multicolumn{1}{b{0.75cm}|}{\cellcolor[HTML]{C9DAF8}{\color[HTML]{6AA84F} \textbf{25}}} &
  \cellcolor[HTML]{C9DAF8}47 &
  \multicolumn{1}{b{0.75cm}|}{\cellcolor[HTML]{C9DAF8}{\color[HTML]{6AA84F} 37}} &
  \multicolumn{1}{b{0.75cm}|}{\cellcolor[HTML]{C9DAF8}{\color[HTML]{FF0000} \underline{\textbf{63}}}}&
  \cellcolor[HTML]{C9DAF8}0 &
  \multicolumn{1}{b{0.75cm}|}{\cellcolor[HTML]{C9DAF8}{\color[HTML]{6AA84F} 70}} &
  \multicolumn{1}{b{0.75cm}|}{\cellcolor[HTML]{C9DAF8}{\color[HTML]{FF0000} \textbf{30}}} &
  \cellcolor[HTML]{C9DAF8}{\color[HTML]{FF0000} 0} &
  \multicolumn{1}{b{0.75cm}|}{\cellcolor[HTML]{C9DAF8}{\color[HTML]{FF0000} 58.39}} &
  \multicolumn{1}{b{0.75cm}|}{\cellcolor[HTML]{C9DAF8}{\color[HTML]{6AA84F} \textbf{29.35}}} &
  \cellcolor[HTML]{C9DAF8}{\color[HTML]{6AA84F} 12.26} \\ \hline
 &
  \cellcolor[HTML]{C9DAF8}\textbf{P} &
  \multicolumn{1}{b{0.75cm}|}{\cellcolor[HTML]{C9DAF8}{\color[HTML]{FF0000} 0}} &
  \multicolumn{1}{b{0.75cm}|}{\cellcolor[HTML]{C9DAF8}{\color[HTML]{6AA84F} \textbf{100}}} &
  \cellcolor[HTML]{C9DAF8}0 &
  \multicolumn{1}{b{0.75cm}|}{\cellcolor[HTML]{C9DAF8}{\color[HTML]{6AA84F} 51.00}} &
  \multicolumn{1}{b{0.75cm}|}{\cellcolor[HTML]{C9DAF8}{\color[HTML]{FF0000} \textbf{41.00}}} &
  \cellcolor[HTML]{C9DAF8}{\color[HTML]{6AA84F} 8.00} &
  \multicolumn{1}{b{0.75cm}|}{\cellcolor[HTML]{C9DAF8}{\color[HTML]{6AA84F} 56.65}} &
  \multicolumn{1}{b{0.75cm}|}{\cellcolor[HTML]{C9DAF8}{\color[HTML]{6AA84F} \textbf{40.04}}} &
  \cellcolor[HTML]{C9DAF8}{\color[HTML]{FF0000} 3.31} &
  \multicolumn{1}{b{0.75cm}|}{\cellcolor[HTML]{C9DAF8}{\color[HTML]{FF0000} 4}} &
  \multicolumn{1}{b{0.75cm}|}{\cellcolor[HTML]{C9DAF8}{\color[HTML]{FF0000} \textbf{49}}} &
  \cellcolor[HTML]{C9DAF8}{\color[HTML]{6AA84F} 47} &
  \multicolumn{1}{b{0.75cm}|}{\cellcolor[HTML]{C9DAF8}{\color[HTML]{FF0000} 55}} &
  \multicolumn{1}{b{0.75cm}|}{\cellcolor[HTML]{C9DAF8}{\color[HTML]{6AA84F} \textbf{45}}} &
  \cellcolor[HTML]{C9DAF8}0 &
  \multicolumn{1}{b{0.75cm}|}{\cellcolor[HTML]{C9DAF8}{\color[HTML]{FF0000} 97}} &
  \multicolumn{1}{b{0.75cm}|}{\cellcolor[HTML]{C9DAF8}{\color[HTML]{6AA84F} \textbf{3}}} &
  \cellcolor[HTML]{C9DAF8}0 &
  \multicolumn{1}{b{0.75cm}|}{\cellcolor[HTML]{C9DAF8}{\color[HTML]{FF0000} 29.74}} &
  \multicolumn{1}{b{0.75cm}|}{\cellcolor[HTML]{C9DAF8}{\color[HTML]{6AA84F} \underline{\textbf{64.74}}}} &
  \cellcolor[HTML]{C9DAF8}{\color[HTML]{FF0000} 5.52} \\ \cline{2-23} 
 &
  \cellcolor[HTML]{C9DAF8}\textbf{MP} &
  \multicolumn{1}{b{0.75cm}|}{\cellcolor[HTML]{C9DAF8}{\color[HTML]{FF0000} 1}} &
  \multicolumn{1}{b{0.75cm}|}{\cellcolor[HTML]{C9DAF8}{\color[HTML]{6AA84F} \textbf{99}}} &
  \cellcolor[HTML]{C9DAF8}{\color[HTML]{6AA84F} 1} &
  \multicolumn{1}{b{0.75cm}|}{\cellcolor[HTML]{C9DAF8}{\color[HTML]{6AA84F} 28.83}} &
  \multicolumn{1}{b{0.75cm}|}{\cellcolor[HTML]{C9DAF8}{\color[HTML]{FF0000} \textbf{27.26}}} &
  \cellcolor[HTML]{C9DAF8}{\color[HTML]{6AA84F} 43.91} &
  \multicolumn{1}{b{0.75cm}|}{\cellcolor[HTML]{C9DAF8}{\color[HTML]{6AA84F} 60.17}} &
  \multicolumn{1}{b{0.75cm}|}{\cellcolor[HTML]{C9DAF8}{\color[HTML]{FF0000} \textbf{30.17}}} &
  \cellcolor[HTML]{C9DAF8}{\color[HTML]{6AA84F} 9.66} &
  \multicolumn{1}{b{0.75cm}|}{\cellcolor[HTML]{C9DAF8}{\color[HTML]{FF0000} 13}} &
  \multicolumn{1}{b{0.75cm}|}{\cellcolor[HTML]{C9DAF8}{\color[HTML]{6AA84F} \textbf{37}}} &
  \cellcolor[HTML]{C9DAF8}{\color[HTML]{6AA84F} 50} &
  \multicolumn{1}{b{0.75cm}|}{\cellcolor[HTML]{C9DAF8}{\color[HTML]{FF0000} 16}} &
  \multicolumn{1}{b{0.75cm}|}{\cellcolor[HTML]{C9DAF8}{\color[HTML]{6AA84F} \underline{\textbf{84}}}} &
  \cellcolor[HTML]{C9DAF8}0 &
  \multicolumn{1}{b{0.75cm}|}{\cellcolor[HTML]{C9DAF8}{\color[HTML]{FF0000} 44}} &
  \multicolumn{1}{b{0.75cm}|}{\cellcolor[HTML]{C9DAF8}{\color[HTML]{6AA84F} \textbf{56}}} &
  \cellcolor[HTML]{C9DAF8}0 &
  \multicolumn{1}{b{0.75cm}|}{\cellcolor[HTML]{C9DAF8}{\color[HTML]{FF0000} 34.35}} &
  \multicolumn{1}{b{0.75cm}|}{\cellcolor[HTML]{C9DAF8}{\color[HTML]{FF0000} \textbf{52.35}}} &
  \cellcolor[HTML]{C9DAF8}{\color[HTML]{6AA84F} 13.30} \\ \cline{2-23} 
\multirow{-3}{*}{\textbf{NF}} &
  \cellcolor[HTML]{C9DAF8}\textbf{MN} &
  \multicolumn{1}{b{0.75cm}|}{\cellcolor[HTML]{C9DAF8}{\color[HTML]{FF0000} 0}} &
  \multicolumn{1}{b{0.75cm}|}{\cellcolor[HTML]{C9DAF8}{\color[HTML]{6AA84F} \textbf{100}}} &
  \cellcolor[HTML]{C9DAF8}{\color[HTML]{FF0000} 0} &
  \multicolumn{1}{b{0.75cm}|}{\cellcolor[HTML]{C9DAF8}{\color[HTML]{6AA84F} 30.30}} &
  \multicolumn{1}{b{0.75cm}|}{\cellcolor[HTML]{C9DAF8}{\color[HTML]{FF0000} \textbf{17.43}}} &
  \cellcolor[HTML]{C9DAF8}{\color[HTML]{6AA84F} 52.26} &
  \multicolumn{1}{b{0.75cm}|}{\cellcolor[HTML]{C9DAF8}{\color[HTML]{6AA84F} 52.91}} &
  \multicolumn{1}{b{0.75cm}|}{\cellcolor[HTML]{C9DAF8}{\color[HTML]{FF0000} \textbf{22.35}}} &
  \cellcolor[HTML]{C9DAF8}{\color[HTML]{6AA84F} 24.74} &
  \multicolumn{1}{b{0.75cm}|}{\cellcolor[HTML]{C9DAF8}{\color[HTML]{6AA84F} 15}} &
  \multicolumn{1}{b{0.75cm}|}{\cellcolor[HTML]{C9DAF8}{\color[HTML]{FF0000} \textbf{23}}} &
  \cellcolor[HTML]{C9DAF8}{\color[HTML]{6AA84F} 62} &
  \multicolumn{1}{b{0.75cm}|}{\cellcolor[HTML]{C9DAF8}{\color[HTML]{FF0000} 4}} &
  \multicolumn{1}{b{0.75cm}|}{\cellcolor[HTML]{C9DAF8}{\color[HTML]{6AA84F} \underline{\textbf{94}}}} &
  \cellcolor[HTML]{C9DAF8}{\color[HTML]{FF0000} 2} &
  \multicolumn{1}{b{0.75cm}|}{\cellcolor[HTML]{C9DAF8}{\color[HTML]{FF0000} 23}} &
  \multicolumn{1}{b{0.75cm}|}{\cellcolor[HTML]{C9DAF8}{\color[HTML]{6AA84F} \textbf{69}}} &
  \cellcolor[HTML]{C9DAF8}{\color[HTML]{FF0000} 8} &
  \multicolumn{1}{b{0.75cm}|}{\cellcolor[HTML]{C9DAF8}{\color[HTML]{FF0000} 29.30}} &
  \multicolumn{1}{b{0.75cm}|}{\cellcolor[HTML]{C9DAF8}{\color[HTML]{6AA84F} \textbf{41.96}}} &
  \cellcolor[HTML]{C9DAF8}{\color[HTML]{6AA84F} 28.74} \\ \hline
\end{tabular}

}

\caption{Experiment results across all models. Background indicates the expected label (green - "entailment", blue - "neutral"). The green and red font colors indicate increase or decrease compared to standalone premises. The numbers in \textbf{bold} also indicate \textit{accuracy} (e.g. for uncertainty adverbs accuracy coincides with the percentage of neutrals). The \underline{\textbf{underlined bold}} numbers indicate \textit{the best accuracy for a specific inference type} across all models (human baseline excluded). \textbf{SA} - standalone premises; \textbf{PT} - premises under presupposition triggers; \textbf{NF} - premises under non-factives; \textbf{P} - pronouns (grammatically-specified-entailments); \textbf{MP} - monotonicity entailment (positive); \textbf{MN} - monotonicity entailment(negative), \textbf{UA} - uncertainty adverbs; \textbf{e} - entailment (\%); \textbf{n} - neutral (\%); \textbf{c} - contradiction (\%).}
\label{tab:all}
\end{table*}
Accuracy across all tested LLMs obtained in the experiments in Section \ref{experiments} is compared in Table \ref{tab:comp}.
\begin{table*}[t!]

\resizebox{\textwidth}{!}{%
\begin{tabular}{|
>{\columncolor[HTML]{FFFFFF}}l |
>{\columncolor[HTML]{FFFFFF}}l |
>{\columncolor[HTML]{FFFFFF}}r |
>{\columncolor[HTML]{FFFFFF}}r |
>{\columncolor[HTML]{FFFFFF}}r |
>{\columncolor[HTML]{FFFFFF}}r |
>{\columncolor[HTML]{FFFFFF}}r |
>{\columncolor[HTML]{FFFFFF}}r |
>{\columncolor[HTML]{FFFFFF}}r |}
\hline
 &
  Inference type &
  \multicolumn{1}{l|}{\cellcolor[HTML]{FFFFFF}human} &
  \multicolumn{1}{l|}{\cellcolor[HTML]{FFFFFF}LLaMA 2 70b} &
  \multicolumn{1}{l|}{\cellcolor[HTML]{FFFFFF}text-davinci-003} &
  \multicolumn{1}{l|}{\cellcolor[HTML]{FFFFFF}gpt-3.5-turbo-0301} &
  \multicolumn{1}{l|}{\cellcolor[HTML]{FFFFFF}gpt-3.5-turbo-0613} &
  \multicolumn{1}{l|}{\cellcolor[HTML]{FFFFFF}gpt-4-0314} &
  \multicolumn{1}{l|}{\cellcolor[HTML]{FFFFFF}gpt-4-0613} \\ \hline
\cellcolor[HTML]{FFFFFF} &
  \textbf{pronouns} &
  \textbf{100.00} &
  31.00 &
  39.00 &
  53.00 &
  9.00 &
  84.00 &
  \textbf{100.00} \\ \cline{2-9} 
\cellcolor[HTML]{FFFFFF} &
  \textbf{monotonicity positive} &
  \textbf{89.00} &
  36.00 &
  25.00 &
  43.00 &
  21.00 &
  39.00 &
  70.00 \\ \cline{2-9} 
\cellcolor[HTML]{FFFFFF} &
  \textbf{monotonicity negative} &
  \textbf{94.00} &
  35.00 &
  28.00 &
  42.00 &
  37.00 &
  88.00 &
  56.00 \\ \cline{2-9} 
\multirow{-4}{*}{\cellcolor[HTML]{FFFFFF}\textbf{Standalone}} &
  \textbf{uncertainty adverbs} &
  {\color[HTML]{374151} \textbf{100.00}} &
  {\color[HTML]{374151} 9.00} &
  {\color[HTML]{374151} 4.67} &
  {\color[HTML]{374151} 9.11} &
  19.00 &
  70.00 &
  38.00 \\ \hline
\cellcolor[HTML]{FFFFFF} &
  \textbf{pronouns} &
  \textbf{100.00} &
  32.00 &
  65.35 &
  72.30 &
  6.00 &
  90.00 &
  \textbf{100.00} \\ \cline{2-9} 
\cellcolor[HTML]{FFFFFF} &
  \textbf{monotonicity positive} &
  \textbf{97.00} &
  36.87 &
  38.04 &
  55.65 &
  17.00 &
  39.00 &
  71.00 \\ \cline{2-9} 
\cellcolor[HTML]{FFFFFF} &
  \textbf{monotonicity negative} &
  \textbf{95.00} &
  38.52 &
  19.04 &
  29.39 &
  24.00 &
  87.00 &
  53.00 \\ \cline{2-9} 
\multirow{-4}{*}{\cellcolor[HTML]{FFFFFF}\textbf{\begin{tabular}[c]{@{}l@{}}Under \\ presupposition \\ triggers\end{tabular}}} &
  \textbf{uncertainty adverbs} &
  \textbf{99.00} &
  29.35 &
  2.65 &
  7.52 &
  25.00 &
  63.00 &
  30.00 \\ \hline
\cellcolor[HTML]{FFFFFF} &
  \textbf{pronouns} &
  \textbf{100.00} &
  64.74 &
  41.00 &
  40.04 &
  49.00 &
  45.00 &
  3.00 \\ \cline{2-9} 
\cellcolor[HTML]{FFFFFF} &
  \textbf{monotonicity positive} &
  \textbf{99.00} &
  52.35 &
  27.26 &
  30.17 &
  37.00 &
  84.00 &
  56.00 \\ \cline{2-9} 
\multirow{-3}{*}{\cellcolor[HTML]{FFFFFF}\textbf{\begin{tabular}[c]{@{}l@{}}Under\\ non-factives\end{tabular}}} &
  \textbf{monotonicity negative} &
  \textbf{100.00} &
  41.96 &
  17.43 &
  22.35 &
  23.00 &
  94.00 &
  69.00 \\ \hline
 &
  \textbf{overall accuracy} &
  \textbf{97.55} &
  36.98 &
  27.95 &
  36.78 &
  24.27 &
  71.18 &
  58.73 \\ \hline
\end{tabular}
}

\caption{Human accuracy and accuracy (\%) across LLMs compared. LLaMA 2, GPT-3.5 (text-davinci-003) and gpt-3.5-turbo-0301 were tested on full data. Human performance, gpt-3.5-turbo-0613 and both GPT-4 versions were tested on 100 examples per inference type.}
\label{tab:comp}
\end{table*}

\section{Prompt Paraphrasing}\label{sec:consistency}
We experimented with the following prompt variations: 
(Original prompt)
\begin{quote}
\textit{You are given a pair of texts. Say about this pair: given Text 1, is Text 2 true, false or neutral (you can't tell if it's true or false)? Reply in one word.}

 \textit{Text 1: ``{text1}''}
 
 \textit{Text 2: ``{text2}'' }
 
\end{quote} 

(Variation 1)
\begin{quote}
\textit{You have two texts, and your task is to determine the truthfulness of Text 2 based on Text 1. Provide a one-word response indicating whether Text 2 is true, false, or neutral (indeterminable). Here are the texts:}

 \textit{Text 1: ``{text1}''}
 
 \textit{Text 2: ``{text2}''}
 
\end{quote} 

(Variation 2)
\begin{quote}
\textit{Assess the veracity of Text 2 based on Text 1: Is Text 2 true, false, or indeterminable? Provide a one-word response.}

 \textit{Text 1: ``{text1}''}
 
 \textit{Text 2: ``{text2}''}
 
\end{quote}

The comparison between the three prompts is shown in Table \ref{tab:prph}.
\begin{table*}[t!]
\resizebox{\textwidth}{!}{%
\begin{tabular}{|l|l|ll|rr|rr|}
\hline
 &
   &
  \multicolumn{2}{l|}{\textbf{original prompt}} &
  \multicolumn{2}{l|}{\textbf{paraphrased prompt 1}} &
  \multicolumn{2}{l|}{\textbf{paraphrased prompt 2}} \\ \hline
 &
   &
  \multicolumn{1}{l|}{\textbf{accuracy (\%)}} &
  \textbf{entailment (\%)} &
  \multicolumn{1}{l|}{\textbf{accuracy (\%)}} &
  \multicolumn{1}{l|}{\textbf{entailment (\%)}} &
  \multicolumn{1}{l|}{\textbf{accuracy (\%)}} &
  \multicolumn{1}{l|}{\textbf{entailment (\%)}} \\ \hline
 &
  \cellcolor[HTML]{D9EAD3}pronouns &
  \multicolumn{1}{l|}{\cellcolor[HTML]{D9EAD3}53.00} &
  \cellcolor[HTML]{D9EAD3}53.00 &
  \multicolumn{1}{r|}{\cellcolor[HTML]{D9EAD3}\textbf{67.00}} &
  \cellcolor[HTML]{D9EAD3}67.00 &
  \multicolumn{1}{r|}{\cellcolor[HTML]{D9EAD3}62.00} &
  \cellcolor[HTML]{D9EAD3}62.00 \\ \cline{2-8} 
 &
  \cellcolor[HTML]{D9EAD3}monotonicity positives &
  \multicolumn{1}{l|}{\cellcolor[HTML]{D9EAD3}43.00} &
  \cellcolor[HTML]{D9EAD3}43.00 &
  \multicolumn{1}{r|}{\cellcolor[HTML]{D9EAD3}\textbf{68.00}} &
  \cellcolor[HTML]{D9EAD3}68.00 &
  \multicolumn{1}{r|}{\cellcolor[HTML]{D9EAD3}65.00} &
  \cellcolor[HTML]{D9EAD3}65.00 \\ \cline{2-8} 
 &
  \cellcolor[HTML]{C9DAF8}monotonicity negatives &
  \multicolumn{1}{l|}{\cellcolor[HTML]{C9DAF8}\textbf{42.00}} &
  \cellcolor[HTML]{C9DAF8}37.00 &
  \multicolumn{1}{r|}{\cellcolor[HTML]{C9DAF8}8.00} &
  \cellcolor[HTML]{C9DAF8}53.00 &
  \multicolumn{1}{r|}{\cellcolor[HTML]{C9DAF8}37.00} &
  \cellcolor[HTML]{C9DAF8}42.00 \\ \cline{2-8} 
\multirow{-4}{*}{\textbf{Standalone}} &
  \cellcolor[HTML]{C9DAF8}uncertainty adverbs &
  \multicolumn{1}{l|}{\cellcolor[HTML]{C9DAF8}{\color[HTML]{374151} \textbf{9.1}}} &
  \cellcolor[HTML]{C9DAF8}{\color[HTML]{374151} 79.9} &
  \multicolumn{1}{r|}{\cellcolor[HTML]{C9DAF8}5.67} &
  \cellcolor[HTML]{C9DAF8}80.33 &
  \multicolumn{1}{r|}{\cellcolor[HTML]{C9DAF8}8.44} &
  \cellcolor[HTML]{C9DAF8}78.67 \\ \hline
 &
  \cellcolor[HTML]{D9EAD3}pronouns &
  \multicolumn{1}{l|}{\cellcolor[HTML]{D9EAD3}72.30} &
  \cellcolor[HTML]{D9EAD3}72.30 &
  \multicolumn{1}{r|}{\cellcolor[HTML]{D9EAD3}\textbf{86.70}} &
  \cellcolor[HTML]{D9EAD3}86.70 &
  \multicolumn{1}{r|}{\cellcolor[HTML]{D9EAD3}76.61} &
  \cellcolor[HTML]{D9EAD3}76.61 \\ \cline{2-8} 
 &
  \cellcolor[HTML]{D9EAD3}monotonicity positives &
  \multicolumn{1}{l|}{\cellcolor[HTML]{D9EAD3}55.65} &
  \cellcolor[HTML]{D9EAD3}55.65 &
  \multicolumn{1}{r|}{\cellcolor[HTML]{D9EAD3}\textbf{78.30}} &
  \cellcolor[HTML]{D9EAD3}78.30 &
  \multicolumn{1}{r|}{\cellcolor[HTML]{D9EAD3}70.39} &
  \cellcolor[HTML]{D9EAD3}70.39 \\ \cline{2-8} 
 &
  \cellcolor[HTML]{C9DAF8}monotonicity negatives &
  \multicolumn{1}{l|}{\cellcolor[HTML]{C9DAF8}29.39} &
  \cellcolor[HTML]{C9DAF8}52.83 &
  \multicolumn{1}{r|}{\cellcolor[HTML]{C9DAF8}6.17} &
  \cellcolor[HTML]{C9DAF8}59.35 &
  \multicolumn{1}{r|}{\cellcolor[HTML]{C9DAF8}\textbf{29.70}} &
  \cellcolor[HTML]{C9DAF8}50.70 \\ \cline{2-8} 
\multirow{-4}{*}{\textbf{\begin{tabular}[c]{@{}l@{}}Under presupposition \\ triggers\end{tabular}}} &
  \cellcolor[HTML]{C9DAF8}uncertainty adverbs (sample) &
  \multicolumn{1}{l|}{\cellcolor[HTML]{C9DAF8}7.52} &
  \cellcolor[HTML]{C9DAF8}86.91 &
  \multicolumn{1}{r|}{\cellcolor[HTML]{C9DAF8}1.39} &
  \cellcolor[HTML]{C9DAF8}88.65 &
  \multicolumn{1}{r|}{\cellcolor[HTML]{C9DAF8}\textbf{9.61}} &
  \cellcolor[HTML]{C9DAF8}82.65 \\ \hline
 &
  \cellcolor[HTML]{C9DAF8}pronouns &
  \multicolumn{1}{l|}{\cellcolor[HTML]{C9DAF8}40.04} &
  \cellcolor[HTML]{C9DAF8}56.65 &
  \multicolumn{1}{r|}{\cellcolor[HTML]{C9DAF8}20.09} &
  \cellcolor[HTML]{C9DAF8}72.87 &
  \multicolumn{1}{r|}{\cellcolor[HTML]{C9DAF8}\textbf{40.52}} &
  \cellcolor[HTML]{F4CCCC}56.65 \\ \cline{2-8} 
 &
  \cellcolor[HTML]{C9DAF8}monotonicity positives &
  \multicolumn{1}{l|}{\cellcolor[HTML]{C9DAF8}\textbf{30.17}} &
  \cellcolor[HTML]{C9DAF8}60.17 &
  \multicolumn{1}{r|}{\cellcolor[HTML]{C9DAF8}9.39} &
  \cellcolor[HTML]{C9DAF8}72.83 &
  \multicolumn{1}{r|}{\cellcolor[HTML]{C9DAF8}29.57} &
  \cellcolor[HTML]{F4CCCC}63.61 \\ \cline{2-8} 
\multirow{-3}{*}{\textbf{\begin{tabular}[c]{@{}l@{}}Under\\ non-factives\end{tabular}}} &
  \cellcolor[HTML]{C9DAF8}monotonicity negatives &
  \multicolumn{1}{l|}{\cellcolor[HTML]{C9DAF8}22.35} &
  \cellcolor[HTML]{C9DAF8}52.91 &
  \multicolumn{1}{r|}{\cellcolor[HTML]{C9DAF8}6.00} &
  \cellcolor[HTML]{F4CCCC}48.87 &
  \multicolumn{1}{r|}{\cellcolor[HTML]{C9DAF8}\textbf{40.00}} &
  \cellcolor[HTML]{F4CCCC}38.39 \\ \hline
\end{tabular}
}
\caption{The experiment results for the original prompt and its two paraphrases suggested by ChatGPT itself. The background color indicates the expected label: green - ``ENTAILMENT'', blue - ``NEUTRAL''. The pink cells indicate the results that do not fit the pattern exhibited by the original prompt (see Appendix \ref{sec:orig_prompt}): predicting more entailment under all embedding types. The bold figures indicate the highest accuracy for a specific inference type across all 3 prompts.}
\label{tab:prph}
\end{table*}

\section{Chain-of-thought prompting}
\label{sec:cot}
\subsection{Chain-of-thought experiments}
For the chain-of-thought experiments, we added the phrase "Let's think step by step" \citep{kojima2023large} to the 
original prompt as follows:

\begin{quote}
\textit{You are given a pair of texts. Say about this pair: given Text 1, is Text 2 true, false or neutral (you can't tell if it's true or false)?}

 \textit{Text 1: ``{text1}''}
 
 \textit{Text 2: ``{text2}''}

\textit{Let's think step by step.}

\end{quote} 

As can be seen, we 1) removed the requirement to return a one-word answer; 2) added the words \textit{``Let's think step by step''} at the end.

After the model outputs a chain of thought, an additional step is needed to obtain a final one-word answer. For this \emph{answer extraction} step we use an additional prompt:

\begin{quote}

\textit{Therefore, the one-word answer (True, False or Neutral) is} 
 
\end{quote} 

For the CoT experiments with standalone premises we use the same 100-example test sets as for the original experiments (see Section \ref{experiments} for details). For experiments with embeddings we sample 100 sentence pairs out of each 2300-example test set.

The results of the original experiments (see Section \ref{experiments}) and the CoT experiments are compared in Table \ref{tab:cot1}.
\begin{table*}[t!]
\resizebox{\textwidth}{!}{%
\begin{tabular}{|l|l|ll|lll|}
\hline
 &
   &
  \multicolumn{2}{l|}{\textbf{original prompt}} &
  \multicolumn{3}{l|}{\textbf{chain-of-though prompt}} \\ \hline
 &
   &
  \multicolumn{1}{l|}{\textbf{accuracy (\%)}} &
  \textbf{entailment (\%)} &
  \multicolumn{1}{l|}{\textbf{accuracy (\%)}} &
  \multicolumn{1}{l|}{\textbf{entailment (\%)}} &
  {\color[HTML]{0000FF} \textbf{neutral (\%)}} \\ \hline
 &
  \cellcolor[HTML]{D9EAD3}\textbf{pronouns} &
  \multicolumn{1}{l|}{\cellcolor[HTML]{D9EAD3}\textbf{53.00}} &
  \cellcolor[HTML]{D9EAD3}53.00 &
  \multicolumn{1}{l|}{\cellcolor[HTML]{D9EAD3}{\color[HTML]{374151} 7.00}} &
  \multicolumn{1}{l|}{\cellcolor[HTML]{D9EAD3}{\color[HTML]{374151} 7.00}} &
  \cellcolor[HTML]{D9EAD3}{\color[HTML]{0000FF} 90.00} \\ \cline{2-7} 
 &
  \cellcolor[HTML]{D9EAD3}\textbf{monotonicity positives} &
  \multicolumn{1}{l|}{\cellcolor[HTML]{D9EAD3}43.00} &
  \cellcolor[HTML]{D9EAD3}43.00 &
  \multicolumn{1}{l|}{\cellcolor[HTML]{D9EAD3}\textbf{44.00}} &
  \multicolumn{1}{l|}{\cellcolor[HTML]{D9EAD3}44.00} &
  \cellcolor[HTML]{D9EAD3}{\color[HTML]{0000FF} 53.00} \\ \cline{2-7} 
 &
  \cellcolor[HTML]{C9DAF8}\textbf{monotonicity negatives} &
  \multicolumn{1}{l|}{\cellcolor[HTML]{C9DAF8}42.00} &
  \cellcolor[HTML]{C9DAF8}37.00 &
  \multicolumn{1}{l|}{\cellcolor[HTML]{C9DAF8}\textbf{53.00}} &
  \multicolumn{1}{l|}{\cellcolor[HTML]{C9DAF8}39.00} &
  \cellcolor[HTML]{C9DAF8}{\color[HTML]{0000FF} 53.00} \\ \cline{2-7} 
\multirow{-4}{*}{\textbf{Standalone}} &
  \cellcolor[HTML]{C9DAF8}\textbf{uncertainty adverbs} &
  \multicolumn{1}{l|}{\cellcolor[HTML]{C9DAF8}{\color[HTML]{374151} 9.1}} &
  \cellcolor[HTML]{C9DAF8}{\color[HTML]{374151} 79.9} &
  \multicolumn{1}{l|}{\cellcolor[HTML]{C9DAF8}\textbf{46.56}} &
  \multicolumn{1}{l|}{\cellcolor[HTML]{C9DAF8}46.44} &
  \cellcolor[HTML]{C9DAF8}{\color[HTML]{0000FF} 46.56} \\ \hline
 &
  \cellcolor[HTML]{D9EAD3}\textbf{pronouns} &
  \multicolumn{1}{l|}{\cellcolor[HTML]{D9EAD3}\textbf{72.30}} &
  \cellcolor[HTML]{D9EAD3}72.30 &
  \multicolumn{1}{l|}{\cellcolor[HTML]{D9EAD3}8.00} &
  \multicolumn{1}{l|}{\cellcolor[HTML]{D9EAD3}8.00} &
  \cellcolor[HTML]{D9EAD3}{\color[HTML]{0000FF} 91.00} \\ \cline{2-7} 
 &
  \cellcolor[HTML]{D9EAD3}\textbf{monotonicity positives} &
  \multicolumn{1}{l|}{\cellcolor[HTML]{D9EAD3}\textbf{55.65}} &
  \cellcolor[HTML]{D9EAD3}55.65 &
  \multicolumn{1}{l|}{\cellcolor[HTML]{D9EAD3}26.00} &
  \multicolumn{1}{l|}{\cellcolor[HTML]{D9EAD3}26.00} &
  \cellcolor[HTML]{D9EAD3}{\color[HTML]{0000FF} 70.00} \\ \cline{2-7} 
 &
  \cellcolor[HTML]{C9DAF8}\textbf{monotonicity negatives} &
  \multicolumn{1}{l|}{\cellcolor[HTML]{C9DAF8}29.39} &
  \cellcolor[HTML]{C9DAF8}52.83 &
  \multicolumn{1}{l|}{\cellcolor[HTML]{C9DAF8}\textbf{56.00}} &
  \multicolumn{1}{l|}{\cellcolor[HTML]{C9DAF8}31.00} &
  \cellcolor[HTML]{C9DAF8}{\color[HTML]{0000FF} 56.00} \\ \cline{2-7} 
\multirow{-4}{*}{\textbf{\begin{tabular}[c]{@{}l@{}}Under presupposition \\ triggers\end{tabular}}} &
  \cellcolor[HTML]{C9DAF8}\textbf{uncertainty adverbs (sample)} &
  \multicolumn{1}{l|}{\cellcolor[HTML]{C9DAF8}7.52} &
  \cellcolor[HTML]{C9DAF8}86.91 &
  \multicolumn{1}{l|}{\cellcolor[HTML]{C9DAF8}\textbf{52.00}} &
  \multicolumn{1}{l|}{\cellcolor[HTML]{C9DAF8}43.00} &
  \cellcolor[HTML]{C9DAF8}{\color[HTML]{0000FF} 52.00} \\ \hline
 &
  \cellcolor[HTML]{C9DAF8}\textbf{pronouns} &
  \multicolumn{1}{l|}{\cellcolor[HTML]{C9DAF8}40.04} &
  \cellcolor[HTML]{C9DAF8}56.65 &
  \multicolumn{1}{l|}{\cellcolor[HTML]{C9DAF8}\textbf{99.00}} &
  \multicolumn{1}{l|}{\cellcolor[HTML]{C9DAF8}1.00} &
  \cellcolor[HTML]{C9DAF8}{\color[HTML]{0000FF} 99.00} \\ \cline{2-7} 
 &
  \cellcolor[HTML]{C9DAF8}\textbf{monotonicity positives} &
  \multicolumn{1}{l|}{\cellcolor[HTML]{C9DAF8}30.17} &
  \cellcolor[HTML]{C9DAF8}60.17 &
  \multicolumn{1}{l|}{\cellcolor[HTML]{C9DAF8}\textbf{78.00}} &
  \multicolumn{1}{l|}{\cellcolor[HTML]{C9DAF8}21.00} &
  \cellcolor[HTML]{C9DAF8}{\color[HTML]{0000FF} 78.00} \\ \cline{2-7} 
\multirow{-3}{*}{\textbf{\begin{tabular}[c]{@{}l@{}}Under\\ non-factives\end{tabular}}} &
  \cellcolor[HTML]{C9DAF8}monotonicity negatives &
  \multicolumn{1}{l|}{\cellcolor[HTML]{C9DAF8}22.35} &
  \cellcolor[HTML]{C9DAF8}52.91 &
  \multicolumn{1}{l|}{\cellcolor[HTML]{C9DAF8}\textbf{58.00}} &
  \multicolumn{1}{l|}{\cellcolor[HTML]{C9DAF8}30.00} &
  \cellcolor[HTML]{C9DAF8}{\color[HTML]{0000FF} 58.00} \\ \hline
\end{tabular}
}
\caption{The experiment results for the original prompt and the CoT prompt. The background color indicates the expected labels: green -- ``ENTAILMENT''; blue -- ``NEUTRAL''. The bold figures indicate which prompt scored higher on a specific inference type.}
\label{tab:cot1}
\end{table*}

\subsection{Manual analysis of ChatGPT's reasoning chains}
\label{cot_analysis}
This expands on the results presented in Section \ref{consistency}, ``Analysis of CoT Results``.
We sampled 55 examples of reasoning chains produced by the ChatGPT at the time of the CoT experiments (5 from each test set) and analyzed them manually.

Some chains of thought (about 14.5\%) are clear, correct, lead to a correct answer and demonstrate understanding of the task and the input premise-hypothesis pair. For example:

\begin{quote}

\textbf{Premise:} \textit{It's worth noting that all philosophers are mortal.}

\textbf{Hypothesis}: \textit{All hegelians are mortal.}

\textbf{Chain of thought}: \textit{Step 1: Examine Text 1. It states that ``All philosophers are mortal.'' This means that every philosopher who ever existed or will exist must eventually die. This statement is categorical and absolute.}

\textit{Step 2: Compare Text 2 to Text 1. Text 2 states that ``All hegelians are mortal.'' Hegelians are a specific subset of philosophers who follow the philosophy of Georg Wilhelm Friedrich Hegel. Since Text 1 states that all philosophers are mortal, and Hegelians fall under the category of philosophers, it follows that Text 2 is true.}

\textit{Answer: True.}

\textbf{Prediction}: \textit{Entailment.}

\end{quote}

In some other cases (about 9.1\%) the prediction is correct, but the reasoning chain is too vague, the logical transitions between the steps are unclear, so that it's impossible to tell if the model reasoned correctly or not. For example:

\begin{quote}

\textbf{Premise:} \textit{Sam violated the school's jewelry policy.}

\textbf{Hypothesis}: \textit{Sam violated something.}

\textbf{Chain of thought}: \textit{1. Start by analyzing Text 1. It states that ``Sam violated the school's jewelry policy.'' This implies that there is a school policy regarding wearing jewelry and Sam broke this rule.}

\textit{2. Next, move on to Text 2. It states that ``Sam violated something.'' It is not specified what Sam violated, but it is implied that Sam did break a rule or a policy of some sort.}

\textit{3. Based on Text 1 and Text 2, it is safe to say that Text 2 is true. The fact that Text 1 explicitly mentions a specific policy that Sam violated supports the idea that a violation did occur. Text 2 confirms this by stating that Sam violated something, thus aligning with the information provided in Text 1.}

\textit{Therefore, the answer is: Text 2 is true.}

\textbf{Prediction}: \textit{Entailment.}

\end{quote}

Yet in other cases (about 23.6\%) the model is ``right for the wrong reasons'' ariving at a correct prediction through a wrong reasoning chain. For example:

\begin{quote}

\textbf{Premise:} \textit{It was reported that every alien used some quarters.}

\textbf{Hypothesis}: \textit{Every alien used coins.}

\textbf{Chain of thought}: \textit{Step 1: 
Let's analyze Text 1 to extract the key information. The statement says ``Every alien used some quarters.'' This means that all aliens used quarters, but it is possible they also used other types of coins or currency.} 

\textit{Step 2: 
Now, let's examine Text 2 in light of Text 1. The statement in Text 2 says ``Every alien used coins.'' This statement seems to suggest that all aliens used coins, but it does not specify which type of coins they used.} 

\textit{Step 3: \textbf{Since Text 2 does not specify which types of coins alien used, we cannot say for sure whether it is true or false}. However, we can say that the statement in Text 2 is neutral since it does not contradict Text 1, but also does not provide us with any new information that supports Text 1.}

\textbf{Prediction}: \textit{Neutral.}

\end{quote}

The correct label here is indeed ``neutral'', however the reason cited in the chain of thought (\textit{``Since Text 2 does not specify which types of coins alien used, we cannot say for sure whether it is true or false''}) is wrong. In reality, the relation is neutral because of the non-factive expression \textit{``It was reported''}.

Remarkably, in one case, the model does not arrive at the gold label, but seems to be - at least pragmatically - correct in its reasoning, and the CoT clearly makes sense:

\begin{quote}

\textbf{Premise:} \textit{it's worth noting that I don't eat chicken skin.}

\textbf{Hypothesis}: \textit{I don't eat chicken skin at any cost.}

\textbf{Chain of thought}: \textit{Step 1: Analyze Text 1
The first text states that the person speaking does not eat chicken skin. This seems pretty clear and straightforward.}

\textit{Step 2: Analyze Text 2
The second text states that the person does not eat chicken skin at any cost. This seems to be an exaggerated statement compared to the first text. The phrase ``at any cost'' implies that the person is willing to go to great lengths or make great sacrifices to avoid eating chicken skin. \textbf{This level of emphasis is not present in Text 1}, so it's possible that Text 2 is an exaggerated version of the truth.}

\textit{Step 3: Decide if Text 2 is true, false, or neutral
Based on the information we have, we can't definitively say whether Text 2 is true or false. It's possible that the person speaking really does avoid eating chicken skin at all costs, but it's also possible that they are exaggerating and simply avoid it when possible. Therefore, the answer is neutral.}

\textbf{Model's prediction}: \textit{Neutral.}

\textbf{Gold label}: \textit{Entailment.}

\end{quote}

Finally, in the remaining cases both the chain of thought and the resulting prediction are wrong. For example:

\begin{quote}

\textbf{Premise:} \textit{The school's deputy headmaster, Howard Jones, told her to remove the crucifix around her neck.}

\textbf{Hypothesis}: \textit{The school's deputy headmaster, Howard Jones, told her to do something.}

\textbf{Chain of thought}: \textit{1. Text 1 states that Howard Jones, the deputy headmaster of the school, asked a student to remove her crucifix.}

\textit{2. Text 2 only confirms that Howard Jones did tell the student to do something, but it doesn't specify what that something was.}

\textit{3. Therefore, Text 2 cannot be used to determine whether or not Text 1 is true or false.}

\textit{4. However, it is important to note that if Text 1 is true, and Howard Jones did ask a student to remove their crucifix, this would be a violation of the student's religious freedom and may constitute discrimination.}

\textbf{Model's prediction}: \textit{Neutral.}

\textbf{Correct prediction}: \textit{Entailment.}

\end{quote}

Here the model not only demonstrates misunderstanding of the task trying to determine the veracity of Text 1 based on Text 2, while clearly instructed to do the opposite, but also makes a moral judgement (see step 4) which is totally out of place in this task.

Below we list some more observations regarding the analyzed reasoning chains:
\begin{enumerate}
 \item The embedding context (presupposition triggers or non-factives) are sometimes mentioned in the chain of thought, but are never used as a basis for the final decision. (One consequence of this is that for the test sets with non-factives all the correct answers result from wrong reasoning chains.)
 \item For premises with uncertainty adverbs, in 8 out of the 10 analyzed cases the adverb is mentioned and its meaning explained, but only in 3 out of the 10 cases the adverb serves as a basis for the final answer. 
 \item The model constantly misinterprets indefinite pronouns as referring to a specific entity. Hence the incorrect ``neutral'' labels for most cases of grammatically-specified entailment. For example ChatGPT decides that \textit{``Mary lent him money''} does not entail \textit{``Someone lent him money''} because ``someone'' in Text 2 \textit{``could be referring to someone other than Mary''}. (The correct answer is, of course, ``entailment'' because ``someone'' is a generic term encompassing \emph{any} individual including Mary.)
 \item The model often gets confused about the monotonicity entailment directions (upward vs. downward), stating, for example, that \textit{``No alien ate pork''} entails \textit{``No alien ate meat''} since \textit{``pork is a type of meat''}, but \textit{``Every alien used some quarters''} does \emph{not} entail \textit{``Every alien used some coins''} because Text 2 \textit{``does not specify which type of coins they used''}.
 \item More generally, the model usually predicts entailment when Text 2 contains a more specific mention than Text 1 (which is, in fact, only correct for cases of downward entailment), and vice versa.
 \item The reasoning chains are mostly vague, excessively wordy, with unclear logical relations between steps, which makes them hard to understand and analyse, and often contain obvious logical errors (e.g. \textit{``Text 2 is likely true, as it directly contradicts the assumption made in Text 1''}).
 \item The CoT can sometimes misrepresent the contents of the input sentences. For example the model claims that the text \textit{``I \underline{love} something outside the city''} does not mention ``love''.
 \item Different chains of thought exhibit contradictory logics. For example, one CoT says \textit{``There is no contradiction between the two texts... Therefore, Text 2 can be determined as \underline{true}''}, while another reasoning chain states: \textit{``Text 2 does not contradict Text 1, so it is \underline{neutral}.''} 
 
\end{enumerate}
Quantitatively, the results of this analysis are represented in Table \ref{tab:cot_analysis}.
\begin{table}[t!]
    \centering
    \resizebox{\columnwidth}{!}{\begin{tabular}{|l|l|}
    \hline
        correct CoT/correct label & 23.6\% \\ \hline
        wrong CoT/correct label & 23.6\% \\ \hline
        wrong CoT/wrong label & 50.9\% \\ \hline
        correct CoT /wrong label & 1.82\% \\ \hline
        CoT coherent and clear & 16.4\% \\ \hline
        underlying LP mentioned in CoT & 49.1\% \\ \hline
        correct understanding of the underlying LP reflected in CoT & 23.6\% \\ \hline
        underlying LP explicitly used in prediction & 14.5\% \\ \hline
        CoT demonstrates correct understanding of the task & 81.8\% \\ \hline
        CoT reflects correct understanding of the input sentences & 80.0\% \\ \hline
    \end{tabular}}
\caption{Manual CoT analysis results. LP stands for "linguistic phenomena". Some numbers are approximate, since not all the cases are clear-cut, and some reasoning chains are unclear and difficult to analyze.}
\label{tab:cot_analysis}
\end{table}

The analysis shows that zero-shot CoT prompting fails to improve ChatGPT's performance on the task because of various flaws in the generated reasoning chains.
 
\end{document}